\pdfoutput=1

\documentclass[11pt]{article}
\usepackage[table]{xcolor}
\definecolor{lightgray}{gray}{0.8}

\usepackage[final]{acl}
\usepackage{colortbl}  %
\usepackage{makecell}  %
\usepackage{booktabs}  %
\usepackage{times}
\usepackage{latexsym}
\usepackage{soul}
\usepackage{tabularx}
\usepackage{fancyhdr}
\usepackage{threeparttable}

\usepackage{multirow, makecell}

\usepackage[utf8]{inputenc}

\usepackage{microtype}

\usepackage[scaled=.9]{inconsolata}
\usepackage{booktabs}
\usepackage{enumitem}

\usepackage{graphicx}

\usepackage{booktabs}
\usepackage{subcaption}
\usepackage{rotating}

\usepackage{amsmath}
\usepackage{multicol}
\usepackage{multirow}

\usepackage{rotating}

\usepackage{soul}

\definecolor{verylightgray}{gray}{0.85}
\definecolor{lightblue}{rgb}{0.7, 0.85, 1}

\newcommand{\hllightgray}[1]{\sethlcolor{verylightgray}\hl{\texttt{#1}}}
\newcommand{\hllightblue}[1]{\sethlcolor{lightblue}\hl{\texttt{#1}}}

\newcommand{\hlgreen}[1]{\sethlcolor{green}\hl{\texttt{#1}}}
\newcommand{\hlblue}[1]{\sethlcolor{cyan}\hl{\texttt{#1}}} %
\newcommand{\hlyellow}[1]{\sethlcolor{yellow}\hl{\texttt{#1}}}

\newcommand{\hlpink}[1]{\sethlcolor{pink}\hl{\texttt{#1}}}
\newcommand{\hlgray}[1]{\sethlcolor{lightgray}\hl{\texttt{#1}}}

\usepackage[normalem]{ulem}

\usepackage{amsfonts}
\usepackage{tcolorbox}

\usepackage{listings}
\usepackage{xcolor}

\usepackage{fancyvrb}
\usepackage{xcolor}
\usepackage{mdframed}
\usepackage{float} 
\usepackage{placeins}
\newmdenv[
    backgroundcolor=gray!20,
    innerleftmargin=10pt,
    innerrightmargin=10pt,
    innertopmargin=10pt,
    innerbottommargin=10pt,
    linewidth=0pt,
  breaklines=true,
    roundcorner=5pt
]{verbatimbox}

\usepackage{mdframed}
\newmdenv[
  backgroundcolor=gray!20,
  linecolor=gray!20,
  innerleftmargin=10pt,
  innerrightmargin=10pt,
  innertopmargin=10pt,
  innerbottommargin=10pt,
  linewidth=8pt,
  roundcorner=16pt,
  keepspaces=true,
  breaklines=true,
  frame=single
]{datasetbox}

\newmdenv[
    backgroundcolor=gray!20,
    innerleftmargin=10pt,
    innerrightmargin=10pt,
    innertopmargin=10pt,
    innerbottommargin=10pt,
    linewidth=0pt,
  keepspaces=true,
  breaklines=true,
    roundcorner=5pt
]{custombox}

\lstdefinestyle{pythonCode}{
  language=Python,
  backgroundcolor=\color{white},
  basicstyle=\footnotesize\ttfamily,
  keywordstyle=\color{blue},
  commentstyle=\color{gray},
  stringstyle=\color{red},
  breaklines=true,
  captionpos=b,
  frame=single,
  showspaces=false,
  showstringspaces=false,
  tabsize=2
}

\lstdefinestyle{phoneBookExample}{
  basicstyle=\ttfamily\small,
  breaklines=true,
  backgroundcolor=\color{gray!20},
  frame=single,
  framerule=0pt,
  framesep=10pt,
  xleftmargin=0pt,
  xrightmargin=0pt,
  keepspaces=true,
  showstringspaces=false
}

\title{Birdie: Advancing State Space Models\\ with Reward-Driven Objectives and Curricula}

\author{
  \textbf{Sam Blouir\textsuperscript{1}},
  \textbf{Jimmy T.H. Smith\textsuperscript{2,3}},
  \textbf{Antonios Anastasopoulos\textsuperscript{1,4}},
  \textbf{Amarda Shehu\textsuperscript{1}}
\\
\\
  \textsuperscript{1}Department of Computer Science, George Mason University, Fairfax, VA \\
  \textsuperscript{2}Stanford University, Stanford, CA \ 
  \textsuperscript{3}Liquid AI, Palo Alto, CA \\
  \textsuperscript{4}Archimedes AI, Athena RC, Athens, Greece\\
    \texttt{\href{mailto:sblouir@gmu.edu}{\{sblouir,antonis,amarda\}@gmu.edu}, \href{mailto:jsmith14@stanford.edu}{jsmith14@stanford.edu}}
}

\newcommand{\cmdFontSize}[2]{\fontsize{#1}{#2}\selectfont}

\begin{document}
\maketitle
\thispagestyle{fancy}

\fancyhf{}  %
\fancyhead[R]{Accepted to EMNLP 2024 (Main Conference)}
\cfoot{\thepage}

\begin{abstract}

Efficient state space models (SSMs), such as linear recurrent neural networks and linear attention variants, offer computational advantages over Transformers but struggle with tasks requiring long-range in-context retrieval-like text copying, associative recall, and question answering over long contexts. Previous efforts to address these challenges have focused on architectural modifications, often reintroducing computational inefficiencies. In this paper, we propose a novel training procedure, Birdie, that significantly enhances the in-context retrieval capabilities of SSMs without altering their architecture. Our approach combines bidirectional input processing with dynamic mixtures of specialized pre-training objectives, optimized via reinforcement learning. We introduce a new bidirectional SSM architecture that seamlessly transitions from bidirectional context processing to causal generation. Experimental evaluations demonstrate that Birdie markedly improves performance on retrieval-intensive tasks such as multi-number phone book lookup, long paragraph question-answering, and infilling. This narrows the performance gap with Transformers, while retaining computational efficiency. Our findings highlight the importance of training procedures in leveraging the fixed-state capacity of SSMs, offering a new direction to advance their capabilities. All code and pre-trained models are available at \url{https://www.github.com/samblouir/birdie}.

\end{abstract}

\begin{figure*}[ht!]
\centering
\noindent\textbf{Figure \ref{fig:phone_mono_chart}: Multi-Phone Number Retrieval}
\begin{tcolorbox}[colframe=black, colback=white, boxrule=1pt, arc=4mm, width=1.9\columnwidth]
    \centering
    \includegraphics[width=\textwidth]{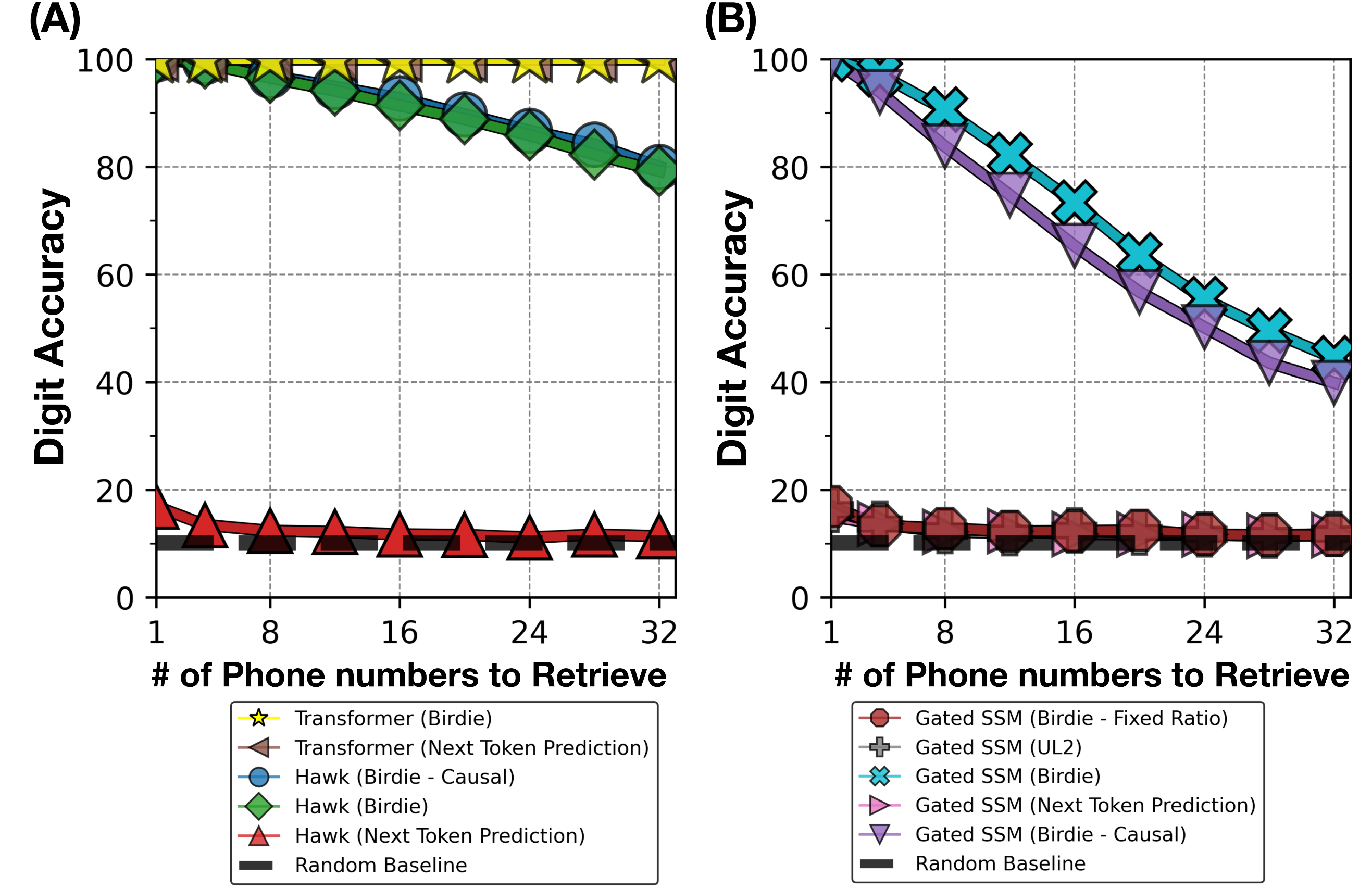}
\end{tcolorbox}
\vspace{-1em}
\caption{
    The \textbf{Multi-Phone Number Retrieval Task} entails finding and retrieving 1-32 phone numbers over a sequence length of 16,384. We demonstrate that State Space Models (SSMs) trained with Birdie significantly reduce their performance disparity with Transformers.
    For further details, please see section \ref{subsec:phonebook}.
    \textbf{(A)} We conduct an ablation study comparing Hawk with Birdie, Birdie - Causal, and Next Token Prediction, alongside a Transformer using Birdie and Next Token Prediction. Hawk trained with Birdie and Birdie - Causal demonstrate significantly higher performance than when trained using Next Token Prediction.
    \textbf{(B)} An ablation that includes UL2 and the Fixed Ratio Mixture on our Gated SSM.
}
\label{fig:phone_mono_chart}
\end{figure*}

\section{Introduction} 
\label{sec:introduction}

Due to their scaling properties~\citep{hoffmann2022training} and in-context learning~\citep{garg2023transformers}, large Transformer models using attention~\citep{bahdanau2014neural, vaswani2017attention} are now prominent in natural language processing (NLP) and achieve effective performance in natural language generation tasks (NLG), including language modeling, machine translation, and question and answering (Q\&A)~\citep{yue2020cliniqg4qa, UnifiedSKG,kumar-etal-21-low}. However,  the softmax attention mechanism cost scales quadratically with sequence length during training, and its key-value (KV) cache grows linearly with sequence length during inference. This leads to increasing costs for training and deployment as model providers continue to increase the context length~\citep{dubey2024llama, reid2024gemini}.

This trend in increasing context length has sparked a strong interest in developing efficient alternative sequence models. The goal is to maintain high performance while scaling effectively with longer sequences. Recent work has focused on recurrent models which offer two key advantages: subquadratic scaling for parallel processing and a fixed state size (in contrast to the growing KV cache in Transformer models) that enables constant-cost inference per step. These models come in different forms, ranging from state space model (SSM)-based methods, such as S4~\citep{gu2022efficiently}, S5~\citep{smith2023simplified}, or Mamba~\citep{gu2023mamba}), to linear RNNs, such as RWKV~\citep{peng2023rwkv}, HGRU~\citep{qin2023hierarchically}, and Hawk~\citep{Griffin-RG-LRU}, to linear attention variants, such as RetNet~\citep{sun2023retentive} and GLA~\citep{yanggated}. These different methods vary in their exact parameterization and parallel computation, but all have an efficient, fixed-state size recurrence for inference. For brevity, we will generally refer to all of these methods as SSMs regardless of their exact parameterization or parallel computation path.

While some studies have shown the ability of SSMs to match Transformers in perplexity and some public benchmarks, an increasing line of work shows that current SSMs struggle on tasks that require long-range in-context abilities~\citep{parkcan}, such as long-range retrieval~\citep{wen2024rnns}, multi-query associative recall~\citep{arora2023zoology, arorasimple}, and copying~\citep{jelassi2024repeat}. These tasks are critical in NLP, where the ability to maintain and manipulate long-term dependencies is key to generating coherent text, following directions, copying sequences, and responding accurately to multiple queries. A typical approach to address these weaknesses has been to formulate hybrid models that interleave SSM layers with global attention layers~\citep{mehtalong,fu2023hungry, parkcan, polimechanistic}, or sliding window attention~\citep{beltagy2020longformer, arorasimple, Griffin-RG-LRU}.\footnote{Sliding window attention, introduced in Longformer~\citep{beltagy2020longformer}, can be viewed as a type of fixed-state size method.} However, models with global attention layers still scale quadratically with sequence length and have a growing KV cache. Models that rely on sliding window attention also fail to perform in-context retrieval outside of the sliding window length~\citep{arorasimple, Griffin-RG-LRU}. 

\vspace{0.5cm}
The predominant focus on architecture to improve performance on long-range in-context abilities misses an opportunity to investigate the role of the pre-training objectives and the potential interaction between the training procedure and model architecture. We note that prior work on generative SSMs exclusively utilizes Next Token Prediction for its pre-training objective. 

In this paper we argue (and show) that in the presence of a fixed state size, a mixture of pre-training objectives can bias learning towards pertinent long-range interactions and that bidirectional processing of the context allows better utilization of the fixed state for such interactions. This paper makes the following key methodological contributions:

\underline{\bf (1)} We develop \textbf{novel pre-training objective mixtures} that confer SSMs strong performance on both standard downstream public benchmarks and recall and copying-intensive tasks where SSMs typically struggle, such as phone book retrieval tasks, infilling, and long paragraph Q\&A. 

\underline{\bf (2)} We show that \textbf{bidirectional processing of the context} combined with the pre-training objective mixtures can further boost performance. In addition, we develop a new bidirectional architecture for SSMs that allows a seamless transition from bidirectional processing of the context to causal generation of the response. 

\underline{\bf (3)} To improve the practical ability to experiment with new pre-training objectives in the mixture, we propose \textbf{a dynamic mixture of pre-training objectives via reinforcement learning (RL)}. This allows for maximizing performance while automating much of the objective selection process. 

The result is a new training procedure that significantly improves the performance of SSMs on recall-intensive tasks, making them more competitive with Transformers. We refer to this procedure as \textit{Birdie}. While we do still observe a performance gap with Transformers on some tasks as the retrieval requirement becomes more difficult (e.g. increasing the number of retrievals required per example), our procedure makes the SSM performance degradation in these scenarios much less severe and expands the regime where these efficient methods can be useful. More broadly, our work points to considering the learning dynamics along with the inductive biases of SSM architectures in order to make better use of the fixed state size.

Our new training procedure, which we call \textit{Birdie}, significantly improves SSM performance on context-heavy and recall-intensive tasks, making this class of models more competitive with Transformers. While a performance gap with Transformers remains on certain tasks as retrieval demands increase (e.g., requiring more retrievals per example), Birdie reduces the severity of performance degradation in these scenarios and extends the range where these efficient models are effective. More broadly, our work underscores the importance of considering both learning dynamics and the inductive biases of SSM architectures to maximize the utility of their fixed state size.

\section{Background and Related Work}
\label{sec:RelatedWork}

This section relates background and prior work.

\subsection{State Space Models} 
Given a length $L$ sequence of inputs $\mathbf{x}_{1:L}\in \mathbb{R}^{L \times D}$, a general class of linear recurrences with hidden states $\mathbf{h}_{1:L}\in \mathbb{R}^{L \times N}$ and outputs $\mathbf{y}_{1:L}\in \mathbb{R}^{L \times D}$ can be computed as shown:
\begin{align*}
    \mathbf{h}_k &= \mathbf{A}_{k}\mathbf{h}_{k-1} + \mathbf{B}_k\mathbf{x}_k\\ 
    \mathbf{y}_k &= \mathbf{g}(\mathbf{h}_k, \mathbf{x}_k)\label{eq:linear_rnn}\\[-2em]
\end{align*}
\vspace*{.5em}
with state transition matrix $\mathbf{A}_{k}\in \mathbb{R}^{N \times N}$, input matrix $\mathbf{B}_{k}\in \mathbb{R}^{N \times U}$ and output function $\mathbf{g}(\cdot)$ to transform the hidden state into an output. 

Many recent recurrent models fall within this SSM framework. Some are time-invariant, such that the dynamics parameters are static across time, i.e. $\mathbf{A}_{k}=\mathbf{A}$ and $\mathbf{B}_{k}=\mathbf{B}\ \forall k$. This includes state space layer/linear RNN variants such as S4~\citep{gu2022efficiently}, S5~\citep{smith2023simplified} and LRU~\citep{LRU-ICML23} and as well as linear attention variants such as linear transformer~\citep{katharopoulos2020transformers} and RetNet~\citep{sun2023retentive}. Other linear recurrent models have input-varying dynamics; these include state space layer/linear RNN variants such as Liquid-S4~\citep{hasani2022liquid}, HGRU~\citep{qin2023hierarchically}, Mamba~\citep{gu2023mamba}, Hawk~\citep{Griffin-RG-LRU}, gated linear attention~\citep{yanggated} methods, and prior work in linear RNNS~\citep{balduzzi2016stronglytyped,martin2018parallelizing,bradbury2016quasirecurrent,lei2018simple}.
The linear (or conditionally linear) dependencies between time steps allow for efficient parallelization across the sequence via Fast Fourier Transforms~\citep{gu2022efficiently, fu2023flashfftconv}, parallel scans~\citep{BlellochTR90, martin2018parallelizing, smith2023simplified} or other structured matrix operations~\citep{yanggated} while also allowing for fast recurrences at inference.

In this work, we focus on input-varying SSMs, as they have provided better performance on language~\citep{gu2023mamba, Griffin-RG-LRU, yanggated} compared to their time-invariant counterparts. This is generally attributed to their ability to ignore or forget contextually-irrelevant information. As an example, consider the Hawk model~\citep{Griffin-RG-LRU} which showed strong performance for attention-free methods on common max-likelihood evaluations. At its core, Hawk is powered by the Real-Gated LRU (RG-LRU), an input-dependent version of LRU. The mathematical formulation of the RG-LRU is:
\vspace{-.5em}
\begin{align*}
    \mathbf{r}_t &= \sigma(\mathbf{W}^{a}\mathbf{x}_t,)\\
    \mathbf{i}_t &= \sigma(\mathbf{W}^{x}\mathbf{x}_t), \\
    \mathbf{a}_t &= \sigma(\Lambda)^{cr_{t}}\\
    \mathbf{h}_t &= \mathbf{a}_t \odot \mathbf{h}_{t-1} + \sqrt{1-\mathbf{a}_t^2} \odot (\mathbf{i}_t \odot \mathbf{x}_t) \\[-2em]
\end{align*}
where $\sigma$ denotes the logistic-sigmoid function, $\Lambda$ is a learnable parameter, and the constant $c$ is set to 8.

\subsection{Weaknesses of Current SSMs}

While the fixed state size allows for efficient deployment at inference time, this limited state capacity also creates a tradeoff in how much information can be stored and used for in-context retrieval. These limitations have been characterized both theoretically~\citep{arora2023zoology, jelassi2024repeat, wen2024rnns} for simple tasks and empirically on both synthetic and more realistic tasks. 

\citet{parkcan} and \citet{arorasimple} show that recurrent models struggle to perform synthetic multi-query associative recall (MQAR)~\citep{arora2023zoology} even when trained directly on the task. \citet{jelassi2024repeat} compared Pythia Transformers~\citep{biderman2023pythia}  to Mamba SSMs~\citep{gu2023mamba} pre-trained on the same dataset and found that Mamba models significantly underperformed the Transformer baselines on retrieval tasks, such as phone-book lookup and long paragraph question-answering. Similarly,~\citet{Griffin-RG-LRU} show that Hawk can perform phone-book retrieval for short lengths but fails to recall the correct phone number as the length grows. In the same work, even the Griffin model, which adds sliding window attention to Hawk struggles to retrieve phone numbers when the task exceeds the sliding window length. This phenomenon is also observed for Based~\citep{arorasimple}, a hybrid of linear attention and sliding window attention on synthetic MQAR tasks. 

Despite their computational appeal, current SSMs display significant weaknesses on the important skill of in-context retrieval. This limits how useful these models can be for practical deployment. We note that these prior works all train models with a simple Next Token Prediction objective. These observations lead us in this work to question the standard training procedure and rethink it as a potential avenue for better utilization of the fixed state size and improved performance on in-context retrieval tasks.

\subsection{Pre-training Objectives}
\label{Pre-trainingObjectiveSection}

Pre-training ``instills'' general-purpose knowledge and abilities~\cite{raffel2020exploring}. While the default choice in NLP for a pre-training objective is Next Token Prediction, several alternative objectives have been proposed that can improve model performance in general language tasks~\citep{Tay2022TranscendingSL, tay2023ul, anil2023palm}, code generation \citep{bavarian2022efficient, rozière2024code}, and multi-modal audio and vision Transformers~\citep{Chen2023PaLIXOS}.

For instance, \emph{Masked Language Modeling} includes objectives where a limited number of tokens are replaced with a mask token, and the model must predict the original tokens. In its original conception with BERT~\citep{devlin2019bert}, each mask token represented one obfuscated input token. \emph{Infilling (Span Corruption)} extends this objective to generative models~\citep{unilm, raffel2020exploring}. For a given input, several spans of tokens are each replaced with unique sentinel tokens. The model then generates the masked tokens and their respective sentinel tokens. In \emph{Prefix language Modeling}, no loss is computed for the prefix, and the model can access the prefix bidirectionally. During pre-training, input sequences are randomly split in two, with the prefix serving as context and the suffix as the target for the direct loss computation~\citep{raffel2020exploring}. The \textit{UL2}~\citep{tay2023ul} objectives combine Prefix language Modeling and Infilling (Span Corruption).

In this paper, we consider and build on the above representative pre-training objectives. As described in Section~\ref{sec:Methods}, we introduce new objectives and \emph{dynamic} mixtures. 

\section{Methods}
\label{sec:Methods}

We propose two key methodological components to reduce the gap between SSMs and Transformers on in-context retrieval tasks: bidirectional processing of the input prompt or prefix and new mixtures of pre-training objectives designed to improve the ability of SSMs to perform retrieval. We then offer a new pre-training procedure that leverages reinforcement learning for dynamic sampling of the pre-training objectives to reduce the burden of pre-selecting the optimal mixture ahead of time. We combine these components to define the \textit{Birdie} training procedure. In the final part of this section, we also describe a baseline Gated SSM that allows for a simple implementation to test our methods.

\subsection{Bidirectional Processing}
\label{subsec:Methods-Bidirec}

Encoder-Decoder models encode a prefix sequence into a single vector (or set of hidden states, for cross-attention) to generate a suffix from~\citep{Kalchbrenner2013_RecurrentTranslationModels, cho2014_encoderDecoderRNN, Sutskever_nmt_2014, raffel2020exploring}. 
The prefix-LM architecture, introduced in T5 and UniLM~\citep{raffel2020exploring, unilm}, simplifies this two-stage approach by instead only using self-attention layers with specific masking to allow for bidirectional processing in the prefix and causality in the suffix regions.
We adapt this prefix-LM approach for recurrent models, which allows us to transfer information between the prefix and suffix regions for every layer, in contrast to the single-vector bottleneck with Encoder-Decoders.

This adaptation ensures our bidirectional SSMs maintain computational and parameter efficiency, enabling fair comparisons with unidirectional models.

To adapt prefix-LM to our models, we first split the recurrent state into forward and reverse components. The forward components are processed without modification, enabling our bidirectional layers to transmit information from the prefix to the suffix via the forward state dimensions. This contrasts with the bidirectional Encoder layers in Encoder-Decoder models, which are constrained to operate only within the prefix. The reverse components are modified in the suffix region to maintain causality; specifically, we zero out the forget gate ($A_t$) dimensions. This approach prevents the state from propagating information backward in causal areas. A mathematical description follows\footnote{Efficient implementations are provided in our codebase: \url{https://www.github.com/samblouir/birdie}}, and an additional example is included in Appendix Section \ref{subsec:Appendix-Bidirectional-Implementation}.

\vspace*{-6mm}
\begin{align*}
    x_{t}^{\text{forward}} &= x_{t, D_{\text{forward}}} \\
    h_{t}^{\text{forward}} &= A_{t} \cdot h_{t-1}^{\text{forward}} + x_{t}^{\text{forward}} \\
    x_{t}^{\text{rev}} &= x_{t, D_{\text{rev}}} \\
    h_{t}^{\text{rev, prefix area}} &= A_{t} \cdot h_{t+1}^{\text{rev}} + x_{t}^{\text{rev}} \\
    h_{t}^{\text{rev, causal area}} &= \textbf{0} \cdot A_{t} \cdot h_{t+1}^{\text{rev}} + x_{t}^{\text{rev}} \\
    h_{t}^{\text{rev}} &= [h_{t}^{\text{rev, prefix area}} \oplus h_{t}^{\text{rev, causal area}}] \\
    h_{t} &= [h_{t}^{\text{forward}} \oplus h_{t}^{\text{rev}}]
\end{align*}

Compared to causal SSMs, our bidirectional SSMs trained with Birdie show greatly enhanced capabilities on question-answering retrieval with information-dense Wikipedia articles in SQuAD V2, shown in Figure~\ref{fig:squad}.
As we allocate half of the state for the reverse direction, we conservatively only allow half of Birdie's layers to be bidirectional. On the multi-phone number retrieval task, we only observe a small reduction
In contexts where the information is Although Birdie - Causal allows for more information to flow forward

Performance is on-par with the Birdie - Causal model in the bandwidth [...]  Our findings demonstrate that these bidirectional SSMs surpass their unidirectional equivalents in multiple such tasks, underscoring the utility of bidirectional processing.

We note that concurrent work, \citet{arora2024just} also applies bidirectional processing to the prefix area with linear attention models and applies independent weights for the prefix and suffix regions. Additionally, they consider a fixed pre-training objective that mixes Prefix language Modeling with Masked Language Modeling, as in work such as UniLM and AlphaCode~\citep{unilm, Li_2022}. Our experimental results with the UL2 objective in Section \ref{Sec:Results} suggest that objectives such as this may be insufficient to improve SSM performance on retrieval-intensive tasks.

 \subsection{Pre-training Objectives for SSMs}
\label{subsec:Methods-Objectives}

We hypothesize that Next Token Prediction does not strongly necessitate in-context retrieval capabilities within SSMs.
For most of the pre-training corpus, much of this objective can be achieved by leveraging information from local tokens alone~\citep{xiao2024duoattention}.
Additionally, common pre-training data preprocessing techniques eliminate repeated or duplicated data in individual training samples~\cite{raffel2020exploring, xue-etal-2021-mt5, olmo}, further reducing the model's need to learn dense copying or long-range retrieval mechanisms.
These factors collectively hinder the model's abilities to retrieve information over long distances.
Although Next Token Prediction does not prevent Transformers from developing long-range retrieval capabilities, SSM architectures inherently possess different inductive biases. 

To enhance the in-context retrieval abilities of SSMs in downstream tasks, we design novel objective mixtures that explicitly train models to learn long-range and high-density retrieval abilities throughout the pre-training process.
We list these objectives and mixtures that we investigate in Table~\ref{table:objective_and_mixture_examples}, and briefly describe several objectives that are core to our new methods:

\textbf{Selective Copying:}
\label{main:selective_copying_text}

We introduce a novel pre-training task, termed Selective Copying, in which the model is trained to retrieve specified spans within a given context, located between designated start and end tokens. An example is provided in Figure \ref{main:selective_copying_example_figure}, and a detailed explanation of the task's format is included in Appendix Section \ref{appendix:selective_copying_example}.
This objective enables SSMs to perform zero-shot text retrieval after pre-training. Figure \ref{fig:APPENDIXRL} in the Appendix demonstrates model performance on this task using validation data from The Pile, which includes sources such as emails and Wikipedia articles. The design of this pre-training objective is inspired by the work of \citet{olsson2022incontext}, which explore similar synthetic induction head tasks.

\begin{figure}[h]
\centering
\noindent\textbf{Figure \ref{main:selective_copying_example_figure}. Selective Copying Example}
\vspace{0.1cm}

\setlength{\tabcolsep}{4pt}  %
\setlength{\extrarowheight}{2pt}  %

\begin{tabular}{|p{\dimexpr 0.9\columnwidth-2\tabcolsep}|} %
\hline
\textbf{Original Input:} \\[-2pt]
~~~~\hllightblue{Birds} \hlpink{sing in} \hlyellow{the} morning. 
\vspace{0.1cm}
\\ \hline

\textbf{Processed Input:} \\[-2pt]
~~~~\texttt{[COPY] [START] \hllightblue{Birds} [END] \hlyellow{the}}  \\
~~~~\texttt{[CONTEXT] \hllightblue{Birds} \hlpink{sing in} \hlyellow{the} morning.} 
\vspace{0.1cm}
\\ \hline

\textbf{Target:} \\[-2pt]
~~~~\texttt{\hlpink{sing in}} 
\vspace{0.1cm}
\\ \hline
\end{tabular}
\caption{
    \textbf{Models retrieve text found between special start and end tokens} in our self-supervised Selective Copying pre-training task. 
    Please see Section \ref{main:selective_copying_text} and Appendix Section \ref{appendix:selective_copying_example} for more details.
    We show model performance on this task in the ``Selective Copying'' column of Figure \ref{fig:APPENDIXRL}.
}
\label{main:selective_copying_example_figure}
\end{figure}

\textbf{Copying:} This is a straightforward recreation of the input sequence, where the model must generate a given input sequence verbatim. This objective is inspired by recent studies \citep{jelassi2024repeat} that discuss the difficulties SSMs face with copying and retrieval tasks.

\textbf{Deshuffling:} The model is presented with a sequence where the tokens are shuffled. The challenge is to reorganize these tokens to restore the original sequence. We implement two variations: one where 50\% of the tokens are shuffled, and another where the entire sequence is shuffled.

\textbf{Autoencoding:} The input sequence is noised using masked spans of tokens, and the model is tasked with generating the original input sequence. This process involves both copying the unmasked tokens from the input and generating new, denoised spans to replace the masked sections.
This approach can be seen akin to BERT's Masked Language Modeling task. However, instead of predicting and infilling masked tokens in their original positions, Autoencoding reconstructs the entire sequence.
An equivalent objective is found in BART's pre-training objectives~\citep{lewis2019bartdenoisingsequencetosequencepretraining}, where the model learns to reconstruct the original text from a corrupted version. Additionally, a similar, albeit masking tokens, rather than spans, strategy is found in T5's ablations~\citep{raffel2020exploring}, referred to as ``Bert-style''.

\textbf{Autoencoding with deshuffling}: This builds on the Autoencoding objective by also shuffling the non-corrupted spans. Effectively, this merges infilling, copying, and de-shuffling into one objective. We hypothesize this may promote transfer learning between objectives. This objective is nearly equivalent to the ``\textit{Text Infilling + Sentence Shuffling}'' objective found in BART's ablations ~\citep{lewis2019bartdenoisingsequencetosequencepretraining}, however, we shuffle sequences between our unmasked spans without any regard to the location of sentence stop words.

\textbf{Fixed Ratio Mixture:} A mixture of all the objectives listed in Table \ref{table:objective_and_mixture_examples} at fixed ratios found by ablation on several downstream tasks using the EleutherAI LM Harness. We provide more details in Appendix Section \ref{Pre-trainingObjectiveSection}. We discuss dynamic ratios in Section \ref{subsec:Methods-Objectives-RL}.

\newcommand{\ObjTableLeftCol}{0.3\columnwidth}
\newcommand{\ObjTableRightCol}{0.65\columnwidth}
\newcommand{\ObjPaddingTop}[0]{\addlinespace[2pt]}
\newcommand{\ObjPaddingBottom}[0]{\addlinespace[2pt]}

\begin{table}[h!tbp]
\centering
\noindent\textbf{Table \ref{table:objective_and_mixture_examples}. Objectives and Mixtures}
\vspace{0.1cm}
\small
\\
\hllightgray{Input: Bird songs fill the early morning air.}
\vspace{0.1cm}
\begin{tabular}{@{}>{\centering\arraybackslash}p{\ObjTableLeftCol}>{\raggedright\arraybackslash}p{\ObjTableRightCol}@{}}
\toprule
\textbf{\cmdFontSize{11}{4}Objectives} & \textbf{\cmdFontSize{11}{4}\hspace{0.05cm}Example} \\
\midrule

\multirow{2}{*}{\makecell[c]{Infilling \\ (Span Corruption)}} & 
\multirow{2}{*}{
\parbox[l]{\linewidth}{
{\small
In: \hspace{0.005cm} \cmdFontSize{8}{4}{Bird [mask] the early [mask]} \\
Tgt: \cmdFontSize{8}{4}{songs fill [mask] air [mask]}
}}} \\ \\
\ObjPaddingBottom
\hline
\ObjPaddingTop

\multirow{2}{*}{\makecell[c]{Next Token \\ Prediction}} & 
\multirow{2}{*}{
\parbox[l]{\linewidth}{
{\small
In: \cmdFontSize{8}{4}{\hspace{0.07cm}Bird songs fill the early morning} \\
Tgt: \cmdFontSize{8}{4}{songs fill the early morning air}
}}} \\ \\
\ObjPaddingBottom
\hline
\ObjPaddingTop

\multirow{2}{*}{\makecell[c]{Prefix Language \\Modeling}} & 
\multirow{2}{*}{
\parbox[l]{\linewidth}{
{\small
In: \cmdFontSize{8}{4}{ Bird songs fill} \\
Tgt: \cmdFontSize{8}{4}{the early morning air}
}}} \\ \\
\ObjPaddingBottom
\hline
\ObjPaddingTop

\multirow{2}{*}{\makecell[c]{Copying}} & 
\multirow{2}{*}{
\parbox[l]{\linewidth}{
{\small
In: \cmdFontSize{8}{4}{ Bird songs fill the early morning air} \\
Tgt: \cmdFontSize{8}{4}{Bird songs fill the early morning air}
}}} \\ \\
\ObjPaddingBottom
\hline
\ObjPaddingTop

\multirow{2}{*}{\makecell[c]{Deshuffling}} & 
\multirow{2}{*}{
\parbox[l]{\linewidth}{
{\small
In: \cmdFontSize{8}{4}{ morning air early fill Bird songs the} \\
Tgt: \cmdFontSize{8}{4}{Bird songs fill the early morning air}
}}} \\ \\
\ObjPaddingBottom
\hline
\ObjPaddingTop

\multirow{2}{*}{\makecell[c]{Autoencoding}} & 
\multirow{2}{*}{
\parbox[l]{\linewidth}{
{\small
In: \cmdFontSize{8}{4}{ Bird [mask] the early [mask]} \\
Tgt: \cmdFontSize{8}{4}{Bird songs fill the early morning air}
}}} \\ \\
\ObjPaddingBottom
\hline
\ObjPaddingTop

\multirow{2}{*}{\makecell[c]{Autoencoding \\ + Deshuffling}} & 
\multirow{2}{*}{
\parbox[l]{\linewidth}{
{\small
In: \cmdFontSize{8}{4}{ the early [mask] Bird [mask]} \\
Tgt: \cmdFontSize{8}{4}{Bird songs fill the early morning air}
}}} \\ \\
\ObjPaddingBottom
\hline
\ObjPaddingTop

\multirow{2}{*}{\makecell[c]{Selective Copying}} & 
\multirow{2}{*}{
\parbox[l]{\linewidth}{
{\small
Please see Figure \ref{main:selective_copying_example_figure} for an example.
}}} \\ \\
\ObjPaddingBottom

\toprule
\textbf{\cmdFontSize{11}{4}Mixtures} & \textbf{\cmdFontSize{11}{4}\hspace{0.05cm}Description} \\
\midrule

\multirow{3}{*}{\makecell[c]{Birdie}} & 
\multirow{3}{*}{
\parbox[l]{\linewidth}{
{\small
Dynamic mixture of above objectives \\
using a reward model for controlling \\
parameterization and sampling ratios. \\
}}} \\ \\ \\
\ObjPaddingBottom
\hline
\ObjPaddingTop

\multirow{3}{*}{\makecell[c]{Fixed Ratio \\Mixture}} & 
\multirow{3}{*}{
\parbox[l]{\linewidth}{
{\small
A mixture of all objectives using a \\
fixed ratio found via ablations on \\
max likelihood tasks. \\
}}} \\ \\ \\
\ObjPaddingBottom
\hline
\ObjPaddingTop
\ObjPaddingTop

\multirow{3}{*}{\makecell[c]{UL2}} & 
\multirow{3}{*}{
\parbox[l]{\linewidth}{
{\small
\cmdFontSize{8}{4}{A Fixed Ratio Mixture consisting of}
\\ \cmdFontSize{8}{4}{Prefix language Modeling and Infilling.}
\\ \cmdFontSize{8}{4}{Described further in Appendix Section \ref{appendix:ul2_explanation}.} \\
}}} \\ \\ \\
\ObjPaddingBottom
\bottomrule
\end{tabular}
\vspace{-1em} 

\caption{This table presents the training objectives and mixtures used in our paper. Birdie effectively parameterizes these objectives, allowing for independent control of multiple factors, a capability that is further elaborated in Appendix Section \ref{appendix:birdie_controls_overview}. Model performance for each objective configuration is detailed in Appendix Section \ref{fig:APPENDIXRL}. `In' denotes the input text; `Tgt' refers to the target output.}

\label{table:objective_and_mixture_examples}
\end{table}

\subsection{Optimal Mixtures with Objective Sampling via Reinforcement Learning}
\label{subsec:Methods-Objectives-RL}

Although we observed promising results in pilot runs, we found it difficult to pre-select optimal task mixture ratios. We also observed that seemingly optimal ratios can change during training, and different model architectures benefit from specialized ratios. Similar challenges in optimally scheduling and adjusting mixtures rates has been noted in~\citet{Tay2022TranscendingSL}.

To address this, we propose a dynamic, automated curriculum that adapts pre-training task mixtures according to the evolving needs of the model. Our approach utilizes a reward model, which we use to predict rewards for proposed actions, given previous actions and observed outcomes. We define actions as training objectives along with their probabilities of being sampled or applied to incoming training data during training.
Overall, this forms a classic multi-armed bandit framework and is related to a recent Gaussian Process approach for dynamic masking rates in Masked Language Modeling ~\cite{urteaga2023multiarmed}, which we found unable to model our diverse objectives and needs. We adopt a four-layer Gated SSM model (See Section \ref{subsec:Methods-GatedRNN}) to directly predict per-objective rewards based on historical training data. We generate random actions and average the top 8 actions with greatest predicted reward.

We visualize loss, greedy-decoding accuracy, and sampling probabilities for training objective categories in Figure~\ref{fig:APPENDIXRL} in Appendix~\ref{sec:appendix}. We observe several trends, such as that training on the Autoencoding objective appears to boost both the Copying and Deshuffling objectives to the extent that their sampling can be nearly shut-off. Other behaviors emerge, such as the selective copying ability continuing to form regardless of sampling rate after the model has seen sufficient amounts of these samples.

We integrate our automated approach with the new objectives introduced in Section~\ref{subsec:Methods-Objectives} and the bidirectional processing from Section~\ref{subsec:Methods-Bidirec} and Appendix~\ref{appendix:sec:Methods-Bidirec}, resulting in a method we call \textit{Birdie}. As shown in Section~\ref{Sec:Results}, Birdie consistently enhances SSM performance across a variety of downstream tasks.

\vspace{4cm}

\subsection{Gated SSM baseline}
\label{subsec:Methods-GatedRNN}

We define a generic Gated SSM baseline to also test our methods on other general SSMs.
\\ The recurrence equations are:
\begin{align*}
    \mathbf{i}_t &= \sigma(\mathbf{W}^{i}\mathbf{x}_t)\in \mathbb{R}^{N}, \\
    \mathbf{z}_t &= \mathbf{W}^{z}\mathbf{x}_t \in \mathbb{R}^{N}, \\ 
    \mathbf{o}_t &= \textsf{GeLU}(\mathbf{W}^{o}\mathbf{x}_t ) \in \mathbb{R}^{N}, \\
    \mathbf{f}_t &= \sigma(\mathbf{W}^{f}\mathbf{x}_t) \in \mathbb{R}^{N}, \\
    \\
    \mathbf{h}_t &= \mathbf{f}_t \odot \mathbf{h}_{t-1} + \mathbf{i}_t \odot \mathbf{z}_t, \\
    \mathbf{y}_t &= \mathbf{W}^{out} (\mathbf{o}_t \odot \mathbf{h}_{t}), %
\end{align*}
where $\sigma$ is the logistic sigmoid function, $\mathbf{x}_t$ is the normalized input at time $t$, and $\mathbf{y}_t$ is the output that feeds into a residual connection. The operator $\odot$ represents element-wise multiplication. We note that this generic Gated SSM is closely related to a parallelizable version of an LSTM~\citep{hochreiter1997long} with the state dependency removed.

In our basic Gated SSM above, we fuse the SSM and MLP blocks as is done in Gated State Spaces and Mamba~\citep{mehtalong, gu2023mamba}. We find this simple baseline performs comparably with state-of-the-art SSMs, such as Hawk, on max-likelihood tasks, but does not perform as well when asked to retrieve multiple phone numbers simultaneously, or when generating responses to questions about Wikipedia excerpts in SQuAD V2.

\section{Experiments and Results}
\label{Sec:Results} 

Here, we present our experimental setup and main findings.

\subsection{Experimental Setup}
\label{subsec:ExperimentalSetup}

We pre-train and instruction-tune a series of 1.4B parameter SSM and Transformer models to investigate the proposed methods. This size allows us to achieve non-trivial performance on popular public benchmarks, while making it feasible to ablate a number of design choices. 

\paragraph{Pre-training:}

We train versions of Hawk, a state-of-the-art SSM, using either Next Token Prediction or the Birdie training procedure described in Section~\ref{subsec:Methods-Objectives-RL}, with its bidirectional prefix processing and dynamic mixture selection. We also include a version without bidirectional prefix processing we refer to as Birdie - Causal.
In addition, we train versions of a modern Transformer architecture using either Next Token Prediction or Birdie\footnote{The Transformer - Birdie variant uses unmasked attention on the prefix, equivalent to the Prefix language Modeling architecture described in \cite{raffel2020exploring}}.

To show our Birdie training procedure is more broadly applicable to other model architectures, we also train several simple 1.4B Gated SSM baseline models, described in Section~\ref{subsec:Methods-GatedRNN} to serve as generalized recurrent models.
In addition, to ablate different aspects of the Birdie design, we train these Gated SSMs using additional objectives and mixtures as described in Table \ref{table:objective_and_mixture_examples}: Next Token Prediction objective, UL2, Fixed Ratio Mixture, Birdie and Birdie - Causal.
Further pre-training details can be found in Appendix~\ref{subsec:Appendix-pre-training}.

\paragraph{Instruction Tuning:} For all models, we loosely follow the progressive learning fine-tuning procedure from Orca 2~\citep{mitra2023orca} and integrate common instruction-tuning procedures from FLAN~\citep{longpre2023flan}, Zephyr~\citep{tunstall2023zephyr}, and Tulu~\citep{wang2023far}. For FLAN, we extend the maximum sequence length to 4096 and further increase it to 8192 for Open-Hermes 2.5~\citep{OpenHermes2.5}. More details on fine-tuning can be found in Sections~\ref{subsec:Appendix-Finetuning}.

\paragraph{Evaluations:}
First, we evaluate our models across 21 tasks using the EleutherAI LM Harness~\citep{eval-harness} to test general knowledge and reasoning abilities and ensure that the Birdie training procedure maintains performance here. We describe these tasks further in Appendix~\ref{subsec:Appendix-EleutherAI} and show per-task performance in Appendix Table~\ref{tab:PerformancePerEleutherAITask}.
We then stress test in-context retrieval abilities of our models by evaluating on tasks previously shown to be difficult for SSMs~\citep{jelassi2024repeat} such as a multi-number phone book lookup task and SQuAD V2 paragraph Q\&A. We also introduce a new infilling dataset to test the models' abilities to comprehend the full context of a story and infill a missing segment.

\begin{figure*}[h]
\centering
\noindent\textbf{Figure \ref{fig:squad}: SQuAD V2 (Question Answering)}
\begin{tcolorbox}[colframe=black, colback=white, boxrule=1pt, arc=4mm, width=2.0\columnwidth]
    \centering
    \includegraphics[width=\columnwidth]{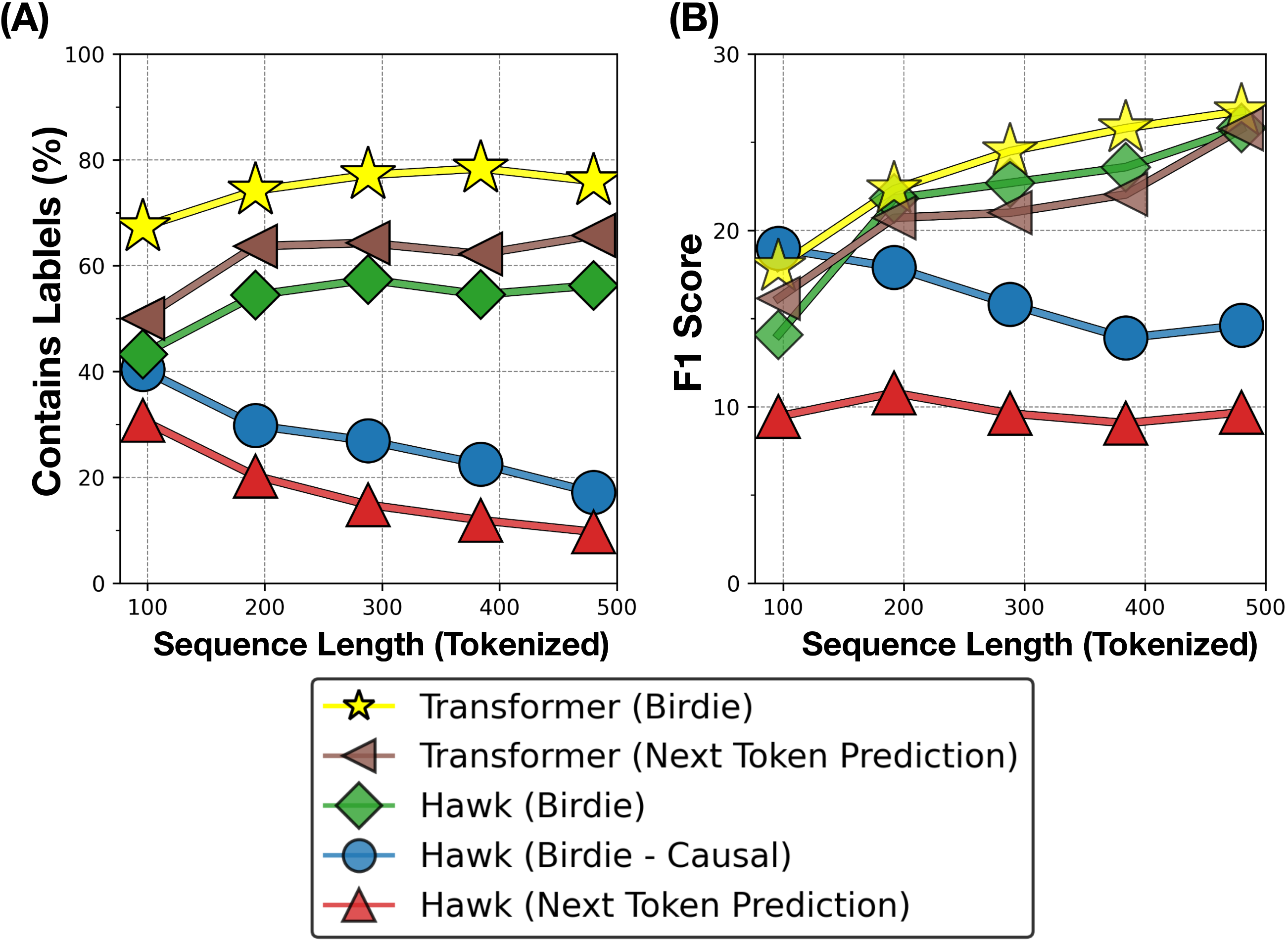}\\[-1mm]
\end{tcolorbox}
\vspace{-1em}
\caption{SQuAD V2 Question-Answering results with instruction-tuned models. Training with the Birdie procedure strongly improves SSM performance, compared to Next Token Prediction. 
Average results are shown in Table \ref{table:squadv2_main_results}.
Further details and ablations are available in Section \ref{sec:Question_Answering_SquadV2} and Appendix Section \ref{appendix:section:squad_v2}.
(A) Answer Contains Label measures when a label is produced by the model verbatim. (B) The F1 Score awards partial credit for matching words in the label and also penalizes models for generating words not in labels. }
\label{fig:squad}
\end{figure*}

\subsection{Comparative Performance and Ablation Study on Maximum-Likelihood Tasks}

We report the average accuracy across 21 unique tasks in Table \ref{avg_eleuther_table}, with specific task-level metrics provided in Appendix Table~\ref{tab:PerformancePerEleutherAITask}.
Our findings indicate that models trained using the Birdie procedure perform comparably to those using the Next Token Prediction objective, demonstrating that Birdie-trained models effectively maintain the knowledge and reasoning skills assessed by these benchmarks.

\begin{table}[h]
\centering

\noindent\textbf{Table \ref{avg_eleuther_table}. Average Accuracy on 21 Tasks}
\vspace{0.1cm}

\begin{subtable}[t]{0.48\textwidth}
    \centering
    \small
    \begin{tabular}{l l r}
    \toprule
    \textbf{Model} & \textbf{Training Procedure} & \textbf{Accuracy (\%)} \\ 
    \midrule
    \multirow{3}{*}{Hawk} 
     & \textbf{Birdie} & \textbf{41.4} \\
     & Birdie (Causal) & 40.8 \\
     & Next Token Prediction & 39.6 \\
    \midrule
    \multirow{2}{*}{Transformer} 
     & Birdie & 39.7 \\
     & \textbf{Next Token Prediction} & \textbf{40.4} \\
    \bottomrule
    \end{tabular}
    \caption{\textbf{Instruct Models}}
    \label{avg_eleuther_instruct_table}
\end{subtable}

\vspace{0.5cm}

\begin{subtable}[t]{0.48\textwidth}
    \centering
    \small
    \begin{tabular}{l l r}
    \toprule
    \textbf{Model} & \textbf{Training Procedure} & \textbf{Accuracy (\%)} \\ 
    \midrule
    \multirow{3}{*}{Hawk} 
     & Birdie & 38.3 \\
     & \textbf{Birdie (Causal)} & \textbf{39.0} \\
     & \textbf{Next Token Prediction} & \textbf{39.1} \\
    \midrule
    \multirow{2}{*}{Transformer} 
     & Birdie & 38.5 \\
     & \textbf{Next Token Prediction} & \textbf{39.9} \\
    \bottomrule
    \end{tabular}
    \caption{\textbf{Base Models}}
    \label{avg_eleuther_base_table}
\end{subtable}

\caption{Average accuracy (\%) on 21 tasks, including ARC, MMLU, and LogiQA. Models trained using Birdie perform comparably to Next Token Prediction. More details and ablations can be found in Appendix section~\ref{tab:PerformancePerEleutherAITask}. (A) Instruction-tuned models fine-tuned on FLAN and OpenHermes 2.5 after pre-training. (B) Base models after pre-training.}
\label{avg_eleuther_table}

\end{table}

\subsection{Analysis on Multi-Phone Number Retrieval} \label{subsec:phonebook}

Next, we explore the phone number retrieval task that previous works have found SSMs trained at various scales struggle on~\citep{jelassi2024repeat, Griffin-RG-LRU, waleffe2024empiricalstudymambabasedlanguage}. In addition, to make it more challenging and further stress-test retrieval capacity, we also measure the simultaneous retrieval of up to 32 numbers from phone books containing 750-800 entries. 

All models underwent minimal fine-tuning from their base configurations, primarily to extend the positional encodings of Transformers to handle longer sequence lengths-from 2,048 to 16,384 tokens. The fine-tuning process spanned 1,000 steps with a batch size of 64, utilizing training samples containing a uniformly sampled number of entries ranging from 200 to 800. For additional details, please see Appendix Section~\ref{appendix:section:phone_number_task_and_sample}.

We summarize the main multi-phone number retrieval results in  Figure \ref{fig:phone_mono_chart}A. We note that, as expected, the Transformer models achieve high performance regardless of the quantity of phone numbers being retrieved, and whether they are trained with either Birdie or Next Token Prediction. However, the Birdie trained Transformer reaches near perfect accuracy much sooner than the Next Token Prediction model.

We observe that SSMs trained with Next Token Prediction perform poorly, even when asked to retrieve only a single phone number and after thorough hyperparameter ablations.  In contrast, we see that SSMs trained with the Birdie procedure significantly outperforms the Next Token Prediction SSM across the regime of different amounts of phone numbers to be retrieved. In fact, the Birdie SSMs achieve 100\% accuracy across 1,024 unique phone books when retrieving a single number and overall significantly reduces the performance gap with the Transformer baselines. While we do observe the performance of the Birdie SSMs degrades as the task complexity increases (i.e. increasing the quantity of phone numbers to be retrieved to 32), the Birdie procedure creates a regime in which SSMs can perform these types of retrievals. We theorize that a larger 7B parameter SSM trained using Birdie may match the Transformer's performance on this task.

Next, we examine the phone number retrieval task, a challenge which previous works have shown SSMs at various scales struggle with~\citep{jelassi2024repeat, Griffin-RG-LRU, waleffe2024empiricalstudymambabasedlanguage}. To increase the task’s difficulty and further stress-test retrieval capacity, we also measure the simultaneous retrieval of up to 32 numbers from phone books containing between 750–800 entries across 16,384 tokens.

All models underwent minimal fine-tuning from their base configurations, mainly to extend the positional encodings of Transformers to handle longer sequence lengths, from 2,048 to 16,384 tokens. The fine-tuning process spanned 1,000 steps with a batch size of 64, using training samples that contained a uniformly sampled number of entries ranging from 200 to 800. For additional details, please refer to Appendix Section~\ref{appendix:section:phone_number_task_and_sample}

We summarize the main multi-phone number retrieval results in Figure \ref{fig:phone_mono_chart}A. As expected, Transformer models achieve high performance regardless of the quantity of phone numbers being retrieved, whether trained with Birdie or Next Token Prediction. However, the Transformer trained with Birdie achieves near-perfect accuracy with far fewer steps needed than the Next Token Prediction model.

Additionally, we observe that SSMs trained with Next Token Prediction perform poorly, even when retrieving a single phone number, despite thorough hyperparameter tuning\footnote{We describe the hyperparameter ablations done to improve non-Birdie models on the multi-phone number retrieval task in Appendix Section~\ref{appendix:section:phone_number_task_and_sample}}. In contrast, SSMs trained with the Birdie procedure significantly outperform the Next Token Prediction SSM across varying retrieval demands. Birdie-trained SSMs achieve 100\% accuracy across 1,024 unique phone books when retrieving a single number, eliminating the performance gap with Transformer baselines. While Birdie-trained SSM performance decreases as task complexity increases (e.g., retrieving up to 32 phone numbers), the Birdie procedure enables SSMs to perform these types of retrievals in the first place. As our models only have 1.4B parameters, we hypothesize that a larger 7B parameter SSM trained with Birdie may be able to match Transformer performance on this task.

\begin{table}[h]
\centering
\noindent\textbf{Table \ref{table:squadv2_main_results}: SQuAD V2 Question-Answering}
\vspace{0.1cm}

\begin{subtable}[t]{0.48\textwidth}
    \centering
    \small
    \begin{tabular}{l l c}
    \toprule
    \multirow{2}{*}{\textbf{Model}} & \multirow{2}{*}{\textbf{Training Procedure}} & \textbf{Contains}\\ & & \textbf{Labels (\%)} \\ 
    \midrule
    \multirow{3}{*}{Hawk} 
     & \textbf{Birdie} & \textbf{55.8} \\
     & Birdie (Causal) & 27.2 \\
     & Next Token Prediction & 16.6 \\
    \midrule
    \multirow{2}{*}{Transformer} 
     & \textbf{Birdie} & \textbf{76.1} \\
     & Next Token Prediction & 63.6 \\
    \bottomrule
    \end{tabular}
    \caption{\textbf{Contains Labels (\%)}}
    \label{squadv2_contains_labels_table}
\end{subtable}

\vspace{0.5cm}

\hfill %
\begin{subtable}[t]{0.48\textwidth}
    \centering
    \small
    \begin{tabular}{l l c}
    \toprule
    \textbf{Model} & \textbf{Training Procedure} & \textbf{F1} \\ 
    \midrule
    \multirow{3}{*}{Hawk} 
     & \textbf{Birdie} & \textbf{22.4} \\
     & Birdie (Causal) & 16.3 \\
     & Next Token Prediction & 10.0 \\
    \midrule
    \multirow{2}{*}{Transformer} 
     & \textbf{Birdie} & \textbf{23.7} \\
     & Next Token Prediction & 21.0 \\
    \bottomrule
    \end{tabular}
    \caption{\textbf{F1 Scores}}
    \label{squadv2_f1_scores_table}
\end{subtable}

\caption{
    Results for instruction-tuned models on SQuAD V2 Question-Answering, where models are given a Wikipedia excerpt and then asked questions about it. This detailed breakdown shows different aspects of model performance. Plots versus context length shown in Figure \ref{fig:squad}. Details and ablations are available in Section \ref{sec:Question_Answering_SquadV2} and Appendix Section \ref{appendix:section:squad_v2}.
    (A) Answer Contains Label measures when a label is produced by the model verbatim.
    (B) The F1 Score awards partial credit for matching words in the label and also penalizes models for generating words not in labels. 
}
\label{table:squadv2_main_results}
\end{table}

\paragraph{Ablations}

We also ablate variations of the Birdie training procedure on this task using the basic Gated SSM model. We show these results in Figure~\ref{fig:phone_mono_chart}B.
The Birdie procedure significantly enhances performance across all tasks compared to all other training procedures considered. We observe a slight but consistent performance boost of the Birdie trained model over the Birdie - Causal trained model, indicating the usefulness of the bidirectional processing of the prefix for the Gated SSM. We also observe the UL2 and Fixed Ratio Mixture procedure (which also uses bidirectional processing of the prefix) does not appear to induce the retrieval abilities necessary for phone number retrieval. In addition, the Fixed Ratio Mixture’s lack of improvement provides evidence of the importance of Birdie's dynamic mixtures for superior training.
These same trends generally hold across the Gated SSM ablations for the infilling task (Appendix Section~\ref{appendix:section:infilling_task_info}) and  SQuAD V2 Question-Answering (Appendix Table~\ref{appendix:section:squad_v2}).

\subsection{Question-Answering}
\label{sec:Question_Answering_SquadV2}

We next evaluate performance on the SQuAD V2 Question-Answering task~\citep{rajpurkar2018know}. Using greedy decoding for up to 128 tokens on all answerable questions, we format inputs as \texttt{``\{context\}\textbackslash n\textbackslash n\{question\}''} without including any few-shot examples. We report an ``Answer Contains Label", where a question is considered correct if any of the labels are found in the response, as well as the classic F1 score. 

Table \ref{table:squadv2_main_results} presents the average results and
Figure \ref{fig:squad} shows the results as a function of sequence length. The performance of Next Token Prediction-trained SSMs strongly degrades with increasing context length, as noted by \citet{jelassi2024repeat}. However, Birdie-trained SSMs maintain performance comparable to Transformers across all available sequence lengths. We note that for this task, unlike the Phone Number Retrieval task in the previous section or the Infilling task in the next section, there is a more meaningful gap between the Birdie and Birdie - Causal trained SSMs. This indicates the bidirectional processing of the prefix may be particularly helpful for this task. 
Further ablations, tables, and details are available in Appendix Section \ref{appendix:section:squad_v2}.

\subsection{Infilling Results}

Finally, we introduce a new infilling task to assess models' capabilities in copying, retrieval, and context comprehension. Models are presented with a story containing 3-7 ordered story entries, one of which is made blank. Models then predict the most appropriate option to fill this blank. As on other tasks, we observe that the Birdie procedure allows the SSM models to perform more closely to the Transformer baselines. The Transformer trained with Birdie also improves its performance.
Table~\ref{infilling_results_chart} relates the main results. More results, details, and an example can be found in Appendix~\ref{appendix:section:infilling_task_info}.

\begin{table}[h]
\centering

\noindent\textbf{Table \ref{infilling_results_chart}: Story Infilling}

\begin{subtable}[t]{0.48\textwidth}
    \centering
    \small
    \begin{tabular}{l l r}
    \toprule
    \textbf{Model} & \textbf{Training Procedure} & \textbf{Accuracy (\%)} \\ 
    \midrule
    \multirow{3}{*}{Hawk} 
     & \textbf{Birdie} & \textbf{42.5} \\
     & Birdie (Causal) & 41.5 \\
     & Next Token Prediction & 33.1 \\
    \midrule
    \multirow{2}{*}{Transformer} 
     & \textbf{Birdie} & \textbf{42.2} \\
     & Next Token Prediction & 41.9 \\
    \bottomrule
    \end{tabular}
    \caption{\textbf{Instruct Models}}
    \label{infilling_instruct_table}
\end{subtable}

\vspace{0.5cm}

\begin{subtable}[t]{0.48\textwidth}
    \centering
    \small
    \begin{tabular}{l l r}
    \toprule
    \textbf{Model} & \textbf{Training Procedure} & \textbf{Accuracy (\%)} \\ 
    \midrule
    \multirow{3}{*}{Hawk} 
     & Birdie & 36.6 \\
     & \textbf{Birdie (Causal)} & \textbf{38.5} \\
     & Next Token Prediction & 29.4 \\
    \midrule
    \multirow{2}{*}{Transformer} 
     & Birdie & 39.8 \\
     & \textbf{Next Token Prediction} & \textbf{40.5} \\
    \bottomrule
    \end{tabular}
    \caption{\textbf{Base Models}}
    \label{infilling_base_table}
\end{subtable}

\caption{
    Average accuracy (\%) on the new infilling dataset, where models complete story segments. Birdie-trained SSMs surpass Next Token Prediction-trained SSMs. For data samples and more, please see Appendix section~\ref{appendix:section:infilling_task_info}. 
    (A) Instruction-tuned models fine-tuned on FLAN and OpenHermes 2.5 after pre-training. 
    (B) Base models after pre-training.
}
\label{infilling_results_chart}

\end{table}

\section{Conclusion}

In this work, we investigated the significant impact of the training procedure on the downstream capabilities of State Space Models (SSMs). While prior research highlighted major weaknesses of SSMs on in-context retrieval tasks, we demonstrated that refining the training process can enhance their performance in these areas. Specifically, we proposed a novel combination of bidirectional processing of the prefix with mixtures of specialized pre-training objectives designed to improve infilling, copying, and handling of long-range dependencies. Additionally, we introduced an RL-based dynamic sampling procedure that adaptively selects optimal objective mixtures throughout training. As a result, the Birdie training procedure strongly improves a model's ability to tackle retrieval-heavy tasks where previous SSM methods have struggled. This finding suggests that, despite the simplicity of Next Token Prediction, this objective may not align optimally with the inductive biases inherent in SSM architectures.

Our work posits that SSMs can achieve enhanced performance through careful selection and design of training objectives, offering a novel pathway for improvement beyond architectural modifications. By showcasing substantial performance gains achievable through this approach, we advocate for a broader reconsideration of how SSMs are developed and optimized. The introduction of Birdie exemplifies the benefits this methodology can bring, pointing toward new directions for future research. We hope that our findings will inspire further exploration of pre-training objectives as a critical factor in advancing SSMs and their application to complex NLP challenges.

\section{Limitations}
It is important to note that these experiments were constrained by an academic budget. While our 1.4B models, trained on 32B tokens, are sufficiently large for specific tasks-such as extracting multiple text spans simultaneously-the scalability of these results with larger models and additional data remains uncertain. Although the 8B Mamba and Mamba-2 models, trained on 3.5 trillion tokens with the Next Token Prediction objective, struggle with tasks like phonebook lookup, which our models appear capable of handling~\citep{waleffe2024empiricalstudymambabasedlanguage}, we did not evaluate whether these larger models could be fine-tuned for such tasks. Initial attempts at a 'second-stage' pre-training with our 1.4B models were also unsuccessful.

The simplicity of the Next Token Prediction objective is difficult to surpass in terms of implementation. In contrast, training setups that employ a mixture of objectives require careful tuning to ensure correct implementation.

Another limitation is the availability of long-context evaluations for LLMs. Tasks that cleanly separate parametric knowledge from true in-context reasoning abilities are scarce~\citep{hsieh2024ruler}. This is especially challenging in realistic question-answering tasks, where the knowledge required may already be memorized from training data. While our long-paragraph question-answering and infilling tasks may be susceptible to this issue, synthetic tasks like phonebook retrieval can more reliably assess in-context reasoning, though their practical relevance is often questioned. Ongoing innovation in long-context evaluation methods is crucial for enhancing language models' long-context capabilities, independent of architecture.

Finally, we observe that SSMs’ performance on retrieval tasks degrades faster than Transformer baselines. We do not claim to have fully solved the retrieval problem, and there are likely other limitations of SSMs that were not captured by the tasks explored in our study.

\section*{Acknowledgments}

This work was supported in part by the National Science Foundation Grant No. 2310113 to Amarda Shehu. Antonios Anastasopoulos is also supported by the NSF under award IIS-2327143. Computations were run on Hopper, a research computing cluster provided by the Office of Research Computing at George Mason University, VA (URL: http://orc.gmu.edu) and on Cloud TPUs provided by Google's TPU Research Cloud (TRC) program. We also thank the anonymous reviewers for their constructive feedback.

\thispagestyle{empty}
\nocite{*}
\bibliography{Blouir-Refs}

\newpage
\clearpage

\appendix

\section{Appendix}
\label{sec:appendix}

\subsection{Reinforcement Learning for Objective Sampling}

We propose Birdie, a reinforcement learning-based approach to dynamically adjust the sampling ratios and configurations (parameters) of multiple training objectives during model training. Our goal is to maximize the overall reduction in loss across various objective classes, under the assumption that each class is equally important to minimize. This method, described below, enables the model to learn which objectives and configurations are most beneficial at different stages of training. Critically, Birdie takes into account the interactions between training objectives.

\subsubsection{Objective Classes}

We train our models using a variety of objectives, which each have several configurations or parameterizations. These objective encourage the model to learn various aspects of language, such as next token prediction, Span Corruption, and sequence reordering, or tasks, like deshuffling and selective copying. The objectives we consider are:

\begin{itemize}
    \item Next Token Prediction
    \item Prefix language Modeling
    \item Selective Copying
    \item Copying
    \item Deshuffling
    \item Infilling (Span Corruption)
    \item Autoencoding with Deshuffling
\end{itemize}

Examples of these can be found in Table \ref{table:objective_and_mixture_examples}, and the configurations are shown in Section \ref{appendix:birdie_controls_overview}.

\subsubsection{Actions and Configurations}

In Birdie's framework, an \textbf{action} is a probability vector representing the sampling frequency for each training objective and configuration. To give Birdie additional control over the training process, we create multiple configurations for each objective class by varying parameters such as context sequence length and masking percentage. Each unique configuration is treated as a separate objective in the action space.

For instance, the \textit{Deshuffling} objective includes configurations with varying percentages of shuffled tokens (50\% or 100\%) and sequence length ranges (between 64--512 or 512--1024 tokens). This allows the model to learn not only which objectives are beneficial to train on, as well as which specific parameter settings are most effective.

\subsubsection{Reward Function}

A reward vector, with elements corresponding to each possible objective parameterization, is calculated based on the change in loss achieved by that configuration. The reward function is designed to:

\begin{itemize}
    \item \textbf{Reduce noise}: Small, insignificant changes in loss are scaled down.
    \item \textbf{Maintain scale}: Rewards are normalized to the range \([-1, 1]\) to stabilize learning. Negative rewards provide an intuitive interface for discouraging undesirable actions.
    \item \textbf{Focus on improvement areas}: Diminishing rewards for already low losses prevent over-focusing on well-performing objectives.
\end{itemize}

The reward for each objective configuration is calculated as:

\[
\Delta L = \frac{\text{loss}_{\text{old}} - \text{loss}_{\text{new}}}{\text{loss}_{\text{old}}}, \quad
S = \sqrt{\text{loss}_{\text{old}} \cdot \text{loss}_{\text{new}}},
\]

\[
\text{Reward} = -r \cdot 100 \cdot \tanh \left( r \cdot S \cdot (\Delta L)^3 \right)
\]
\[
\text{Reward} = \text{clip}(\text{Reward}, -1, 1),
\]

where:

\begin{itemize}
    \item \(\Delta L\) is the normalized change in loss.
    \item \(S\) is the geometric mean of the old and new losses.
    \item \(r\) is a sensitivity hyperparameter (by default, we use Euler's number \(e\)).
\end{itemize}

This function balances improvements across different loss scales. For example, reducing a loss from 4.5 to 4.2 yields a similar reward to reducing a loss from 0.6 to 0.5207, despite the difference in percentage changes ($-7\%$ vs $-13\%$).

To normalize for certain objective classes having greater numbers of configurations than others, we compute the \textit{Class Reward} \(R_c\) for each objective class \(c\) by averaging the rewards of all configurations within that class:

\[
R_c = \frac{1}{N_c} \sum_{i \in c} R_i,
\]

where \(N_c\) is the number of configurations within class \(c\) and \(R_i\) is the reward for configuration \(i\). The \textit{Total Reward} is then the sum of all Class Rewards:

\[
R_{\text{total}} = \sum_{c} R_c.
\]

\subsubsection{Choosing an Action}

At each evaluation iteration, the Birdie's action selection procedure involves the following steps.
Please see the pseudocode in Appendix Figure \ref{appendix:pseudocode_for_sampling_action_from_birdie} for added clarity.

\begin{enumerate}
    \item \textbf{Prepare historical data: losses and action}: At each time step, we have two vectors. One represents the loss across objective configurations, and the other represents the action at that time, or sampling probability for each objective configuration. We stack and concatenate these, creating an input of shape (number of time steps, number of objective configurations * 2)
    \item \textbf{Generate candidate actions}: Since we have prepared our historical data, we now pick an action for our current state. We randomly sample 2,048 probability vectors representing potential actions, and concatenate them with the current loss vector. This creates an input of shape (2048, number of objective configurations * 2).
    
    \item \textbf{Predict rewards}: After concatenating the proposed actions with the current loss, we repeat the 2D array containing our historical losses and actions 2048 times. We concatenate these once more to creates an input of shape (2048, number of time steps+1, number of objective configurations * 2). We then process this using the reward model to obtain our predicted rewards per action.
    \item \textbf{Select top actions}: We choose the 8 actions with the highest predicted rewards.
    \item \textbf{Average actions}: Average these top actions and take this final action until the next evaluation.
\end{enumerate}

Birdie is trained on historical loss, action, and uses observed reward data as labels, enabling it to dynamically estimate optimal actions based on past performance.
Due to the limited amount of training samples, we find the best results by Grokking Birdie, training for 200 steps at every evaluation period rather than reaching a target loss.

Training begins with a warm-up phase where the language model is trained with uniform sampling across all objectives for the first 250 steps. Early evaluations run at 10, 50 and 250 steps to create initial training data for Birdie. Birdie is permanently given control of actions at 250 steps. Further evaluations then occur at steps 500, 1000, 1500, 2000, and every 1000 steps thereafter.

\subsubsection{Reward Model Architecture}

The reward model utilizes a Gated SSM architecture (Section~\ref{subsec:Methods-GatedRNN}) with four layers, each with a hidden size of 256. This model takes as input a sequence of historical losses and actions and predicts the future reward for a given action and current losses. We apply independent RMSNorm layers for the loss and action input dimensions.

\subsection{Birdie's controls}\label{appendix:birdie_controls_overview}
Here, we describe the parameters, or configurations, for our objectives.
For a given objective, its configurations are applied as the superset of available settings. 
We then allow Birdie to adaptively set the sampling probabilities for each configuration independently.
When calculating the total reward, we normalize the reward by scaling each configuration's reward by the number of configurations for that objective class.
A plot of the average performance per class is shown in Appendix Figure \ref{fig:APPENDIXRL}.

\paragraph{Infilling (Span Corruption):} 
\begin{enumerate}
    \item The length of the sequence (between 128-256, 256-512, 512-1024, or 1024-2048 tokens long).
    \item The total percentage of the sequence to be masked (5\%, 15\%, or 50\%).
    \item The mean length of spans (3, 8, or 32 tokens long).
\end{enumerate}

\paragraph{Next Token Prediction and Prefix language Modeling:} 
Due to an implementation limitation, we allow for no controls other than how often these objectives are sampled.

\paragraph{Selective Copying:}
\begin{enumerate}
    \item How many spans to find and copy at once (1, 2, 3, or 4 spans).
    \item How often the context is presented either before or after the START and END segments to find appears. (For example, one style presents a phone book and then asks for a specific person's number. The other style first asks for the person's phone number, and then provides the book.)
    \item The length of the context (between 384-768 or 768-1536 tokens long).
\end{enumerate}

\paragraph{Copying:} 
We allow Birdie to control the length of the sequence to copy, from between 64-256 or 256-2014 tokens to copy.

\paragraph{Deshuffling:} 
\begin{enumerate}
    \item The length of the sequence (between 128-256 or 512-1024 tokens long).
    \item How much of the sequence is shuffled (50\% or 100\% tokens are shuffled).
\end{enumerate}

\paragraph{Autoencoding:} 
\begin{enumerate}
    \item The length of the sequence (between 192-384, 384-768, or 768-1536 tokens long).
    \item The total percentage of the sequence to be masked (15\% or 85\%).
    \item The average length of masked spans for a given sequence (3, 8, or 32 mean span width)
    \item Whether to shuffle the non-masked spans or not.
\end{enumerate}

\clearpage
\newpage

\begin{figure*}[h!]
    \begin{tcolorbox}[colframe=black, colback=white, boxrule=1pt, arc=4mm, width=2.0\columnwidth]
        \centering
\begin{lstlisting}[language=Python,style=pythonCode,basicstyle=\ttfamily\small]
# Pseudocode for sampling an action from Birdie

# Prepare action and loss histories
# action_history: (time_steps, num_configs)
# loss_history:   (time_steps, num_configs)
history = concatenate(action_history, loss_history, axis=-1)
# history.shape: (time_steps, num_configs * 2)

# Generate 2048 proposed actions
# current_losses: (num_configs,)
proposed_actions = random_uniform(2048, num_configs)
proposed_actions /= sum(proposed_actions, axis=-1, keepdims=True)
# proposed_actions.shape: (2048, num_configs)

# Prepare inputs for Birdie
current_losses_expanded = repeat(current_losses[None], (2048, 1))
# current_losses_expanded.shape: (2048, num_configs)
input_actions = concatenate(current_losses_expanded, proposed_actions, axis=-1)
# input_actions.shape: (2048, num_configs * 2)

# Repeat history for batch processing
history_expanded = tile(history[np.newaxis, :, :], (2048, 1, 1))
input_sequence = concatenate(history_expanded, input_actions[:, np.newaxis, :], axis=1)
# input_sequence.shape: (2048, time_steps + 1, num_configs * 2)

# Predict rewards using Birdie
estimated_rewards = Birdie(input_sequence)
# estimated_rewards.shape: (2048, time_steps + 1, num_configs)

# Extract rewards for proposed actions
estimated_rewards = estimated_rewards[:, -1, :]
# Apply per-configuration scaling.
# For each objective's configurations, set the elements of scaling_vector to equal (1/num_configurations_for_the_objective).
# Otherwise, objectives with more configurations are over-represented in the total reward below.
estimated_rewards *= scaling_vector

# Compute total rewards
total_rewards = sum(estimated_rewards, axis=-1)
# Select top actions
top_indices = argsort(total_rewards)[-8:]
top_actions = proposed_actions[top_indices]

# Final action is the average of top actions
final_action = mean(top_actions, axis=0)
\end{lstlisting}
\end{tcolorbox}
\caption{Pseudocode for sampling an action from Birdie.}
\label{appendix:pseudocode_for_sampling_action_from_birdie}
\end{figure*}

\clearpage
\newpage

\subsection{Bidirectional Processing}

\label{appendix:sec:Methods-Bidirec}

\subsubsection{Implementation Details}

To efficiently implement bidirectional processing in SSMs, we adapt the prefix-LM architecture used with Transformers ~\citep{unilm, raffel2020exploring, tay2023ul2} to create a simple mechanism that enables bidirectionality in the prefix (inputs) while enforcing causality in the suffix (outputs). We use a careful construction of the input sequences and corresponding masks, shown below.
Assuming masked sequence packing for efficient training, our approach is compute-matched with a causal scan operation.

\paragraph{Input Sequence Processing:}

Consider an example where we have the original input tokens $\{4, 5, 6\}$ and corresponding labels $\{7, 8, 9\}$. We construct a teacher-forced input to the model by concatenating the inputs and labels, with a special token (in our case, 1) inserted to indicate the beginning of the generation phase. These processed inputs become $\{4, 5, 6, 1, 7, 8\}$. The processed labels to calculate the loss on are $\{-, -, -, 7, 8, 9\}$, where the hyphens indicate positions without any loss (i.e., the model is not trained to predict these tokens).

\paragraph{Reset Mask for Sequence Packing and Bidirectionality:}

When packing samples into our training sequences, we reset the SSM's state to block the model from mixing unrelated samples. We do this by creating a "reset mask" that marks the start of each new sample. At these marks, we reset the state to 0.
To manage the reverse flow of information in our SSMs, we re-use the same reset mask used for sequence packing to control the state information flow, in both reverse and forward directions. Extending on the example given above, the reset mask is $\{1, 0, 0, 2, 2, 2\}$, where the value '2' indicates positions where the reverse state components are forcibly reset to enforce causality in the suffix region, and 1 represents where to reset the state as a new sample begins.

\vspace{6.0cm}

\paragraph{State Partitioning and Concatenation:}

In our SSMs, we then partition the state dimensions into forward and reverse components. Let $f_t$ represent the state at time step $t$, with a total dimension of $D_{\text{state}}$. We split $f_t$ into two halves, shown below using NumPy's syntax:

\[
f_t^{\text{forward}} = f_t[..., :D_{\text{state}}/2]
\]
\[
f_t^{\text{reverse}} = f_t[..., D_{\text{state}}/2:]
\]

Similarly, we split the input $x_t$ into forward and reverse components:

\[
x_t^{\text{forward}} = x_t[..., :D_{\text{state}}/2]
\]
\[
x_t^{\text{reverse}} = x_t[..., D_{\text{state}}/2:]
\]

We then apply the reset mask to the forward and reverse state components as shown below, to prevent backward information flow and prevent inter-sample state interference:

\[
f_t^{\text{forward}} = \begin{cases}
    f_t^{\text{forward}} & \text{if } \text{reset\_mask}[t] \neq 1 \\
    0 & \text{if } \text{reset\_mask}[t] = 1
\end{cases}
\]

\[
f_t^{\text{reverse}} = \begin{cases}
    f_t^{\text{reverse}} & \text{if } \text{reset\_mask}[t] \neq 2 \\
    0 & \text{if } \text{reset\_mask}[t] = 2
\end{cases}
\]

With the masked reverse state components, we proceed to compute the forward and reverse recurrent states using the recurrence functions:

\[
h_t^{\text{forward}} = \text{RecurrenceForward}(f_t^{\text{forward}}, x_t^{\text{forward}})
\]
\[
h_t^{\text{reverse}} = \text{RecurrenceReverse}(f_t^{\text{reverse}}, x_t^{\text{reverse}})
\]

After processing the recurrences, we concatenate the forward and reverse recurrent states along the state dimension to obtain the complete state at time step $t$:

\[
h_t = [h_t^{\text{forward}} \oplus h_t^{\text{reverse}}]
\]

As we segmented the state into forward and reverse portions earlier, this final, concatenated h\_t is equivalently sized to a state that would have resulted from using the same SSM fully causally, allowing our parameter count to remain the same.
Additionally, since an equal number of state dimensions are traveling through the sequence, this state segmentation also allows us to compute match our the causal models.
Empirically, we find this bidirectional approach provides benefits even when compute and parameter matched.

\paragraph{State Utilization:}

By utilizing the reset mask to partitioning the state in this manner, we ensure that bidirectionality is available in the prefix region while maintaining causality in the suffix region, as well as preventing interference among sequence packed samples. This bidirectional encoding of the input sequence can enhance the ability of the SSMs to handle varied inputs without violating the causal constraints necessary for generation, with only a minor reduction in state components traveling forwards.
In contrast to an Encoder-Decoder setup, which restricts bidirectional layers to only process tokens in the prefix area, in our bidirectional layers, $f_t^{\text{forward}}$ runs along the entire sequence, just as a standard causal recurrent layer does.

\subsection{Bidirectional Python Implementation Example:}
\label{subsec:Appendix-Bidirectional-Implementation}

We provide an efficient implementation of our approach on our Github page in PyTorch and JAX at \url{https://www.github.com/samblouir/birdie}.
To further clarify our approach, consider the following pseudocode using Python:

\begin{lstlisting}[language=Python,style=pythonCode, breaklines=true]
Prefix language Modeling example:

This enables bidirectionality on the inputs/context, and enforces causality on the labels.

Assuming masked sequence packing for efficient training, this approach is compute-matched with a causal scan operation.

Example:
# We want to prepare a reset_mask for a 
# given input and label

Original inputs: [4, 5, 6]
Original labels: [7, 8, 9]

# We concatenate these for 
# our decoder-only models.
Processed inputs: [4, 5, 6, 1, 7, 8]
Processed labels: [-, -, -, 7, 8, 9] 
# (1 acts as the "begin generating" token.)

# Locations with "2" mark where
# to block state information flow
#  from the right/reverse-direction
Processed reset_mask: [0, 0, 0, 2, 2, 2]

# (For reference, here is a 
# standard Attention mask for 
# a Prefix language Modeling Transformer.
# This designates the 
# bidirectional/encoder area.)
Processed attn_mask: [1, 1, 1, 0, 0, 0] 

# We now can use the reset_mask in our model.
\end{lstlisting}

\vspace{22em}
\begin{lstlisting}[language=Python,style=pythonCode, breaklines=true]
Equivalent abbreviated SSM code:

# First, we partition f_t into two halves.
split_location = (state_size // 2)

# The shape of f_t is (batch size, length, state_dims)
f_t_forward = f_t[..., :split_location]
f_t_reverse = f_t[..., split_location:]

# The shapes of f_t_forward and f_t_reverse are:
#   (batch size, length, state_dims//2)

# We also split x_t (which is (i_t * z_t))
x_t_forward = x_t[..., :split_location]
x_t_reverse = x_t[..., split_location:]

# The shapes of x_t_forward and x_t_reverse are:
#   (batch size, length, state_dims//2)

# Now we use our reset_mask to mask f_t_reverse in causal areas.
f_t_reverse = np.where(
                reverse_mask == 2,
                0,
                f_t_reverse,
              )

# We can then run the recurrence as usual.
h_t_fwd = \
    recurrence_func(
        f_t_forward,
        x_t_forward,
    )
h_t_rev = \
    reverse_recurrence_func(
        f_t_reverse,
        x_t_reverse,
    )

# Finally, we concatenate along the last axis
h_t = concatenate(xs_fwd, xs_rev)

\end{lstlisting}

\clearpage
\subsection{Selective Copying}
\label{appendix:selective_copying_example}

\subsubsection{Example Illustration}

Consider the input sequence ``\texttt{ABCDEF}''.
We use the following variables with randomly selected values:
\begin{itemize}
    \item \textbf{Selected Span:} ``DE''
    \item \textbf{Start Delimiter Length:} ``2''
    \item \textbf{End Delimiter Length:} ``1''
\end{itemize}

These arguments result in the following selected inputs and labels for the model:
\\
\textbf{Processed Input:}
\[
    \small{[\texttt{CONTEXT}] \; \texttt{ABCDEF} \; [\texttt{COPY}] \; [\texttt{START}] \; BC \; [\texttt{END}] \; F}
\]

\textbf{Processed Label:}
\[
    \small{\text{DE} \; [\text{DONE}]}
\]

\subsubsection{Detailed Instructions}

To construct a Selective Copying instance involving a single span, follow the procedure outlined below:

\begin{enumerate}
    \item \textbf{Sample Loading}: 
    Load an input string from the dataset and tokenize it. This tokenized string is referred to as the ``context.'' The model will extract one or more spans from this context.
    
    \item \textbf{Span Selection}: 
    Randomly select at least one contiguous span from the context, with a length between 4 to 32 tokens. If multiple spans are selected, ensure they do not overlap.
    
    \item \textbf{Delimiter Identification}: 
    For each selected span, randomly determine the lengths of the start and end delimiters (ranging from 1 to 8 tokens). Extract the specified number of tokens immediately before the span as the start delimiter and the specified number of tokens immediately after the span as the end delimiter.
    
    \item \textbf{Formatting the Span and Delimiters}: 
    Concatenate the delimiters with the following tokens:
    \[
    \hlgreen{[START]} \; \texttt{<start\_delim>} \; \hlblue{[END]} \; \texttt{<end\_delim>}
    \]
    Prepend this sequence with the \hlyellow{[COPY]} token to indicate a copying task:
    \[
    \hlyellow{[COPY]} \; \hlgreen{[START]} \; \texttt{<start\_delim>} \; \hlblue{[END]} \; \texttt{<end\_delim>}
    \]
    
    \item \textbf{Concatenating the Context}: 
    Prepend the context with the \hlgray{[CONTEXT]} token:
    \[
    \hlgray{[CONTEXT]} \; \texttt{ABCDEF}
    \]
    Combine this with the formatted delimiters either by prepending or appending:
    \begin{itemize}
        \item \textbf{Prepend:}
        \[
        \hlyellow{[COPY]} \; \hlgreen{[START]} \; \texttt{BC} \; \hlblue{[END]} \; \text{F} \; \hlgray{[CONTEXT]} \; \text{ABCDEF}
        \]
        
        \item \textbf{Append:}
        \[
        \hlgray{[CONTEXT]} \; \text{ABCDEF} \; \hlyellow{[COPY]} \; \hlgreen{[START]} \; \texttt{BC} \; \hlblue{[END]} \; \text{F}
        \]
    \end{itemize}
    
    \item \textbf{Sampling Strategy}: 
    The control system, \texttt{Birdie}, determines the frequency of prepending or appending the delimiters, the number of spans to selectively copy, and sets the maximum length of the context.
\end{enumerate}

\subsubsection{Detailed Example}

Let's revisit the sequence ``ABCDEF'' with the span ``DE'' selected:
\begin{itemize}
    \item \textbf{Start Delimiter Length}: 2 tokens (\texttt{BC})
    \item \textbf{End Delimiter Length}: 1 token (\texttt{F})
\end{itemize}

\textbf{Formatted Delimiters:}
\[
\hlyellow{[COPY]} \; \hlgreen{[START]} \; \text{BC} \; \hlblue{[END]} \; \text{F}
\]

\textbf{Final Concatenated Input (Prepend Example):}
\[
\hlyellow{[COPY]} \; \hlgreen{[START]} \; \text{BC} \; \hlblue{[END]} \; \text{F} \; \hlgray{[CONTEXT]} \; \text{ABCDEF}
\]

\textbf{Final Concatenated Input (Append Example):}
\[
\hlgray{[CONTEXT]} \; \text{ABCDEF} \; \hlyellow{[COPY]} \; \hlgreen{[START]} \; \text{BC} \; \hlblue{[END]} \; \text{F}
\]

\clearpage
\subsection{UL2} \label{appendix:ul2_explanation}

\begin{table*}[htpb]
\centering
\begin{tabular}{|c|c|c|}
\hline
\textbf{Paradigm Token} & \textbf{Mean Span Width} & \textbf{Masked Input \%} \\
\hline
\multirow{2}{*}{"R" Denoisers} & 3 tokens & 15\% \\
                               & 8 tokens & 15\% \\
\hline
\multirow{4}{*}{"X" Denoisers} & 3 tokens & 50\% \\
                               & 8 tokens & 50\% \\
                               & 64 tokens & 15\% \\
                               & 64 tokens & 50\% \\
\hline
"S" Denoisers & \multicolumn{2}{c|}{Prefix language Modeling, sequence split 75\%} \\
\hline
\end{tabular}
\caption{Summary of Denoising Objectives Used in UL2}
\label{appendix:tab:ul2_denoisers_table}
\end{table*}

\begin{table*}[htpb]
\centering
\begin{tabular}{|c|c|c|}
\hline
\textbf{Paradigm Token} & \textbf{Mean Span Width} & \textbf{Masked Input \%} \\
\hline
\multirow{1}{*}{"R" dDenoiser} & 3 tokens & 15\% \\
\hline
\multirow{1}{*}{"X" dDenoiser} & 32 tokens & 50\% \\
\hline
"S" dDenoiser & \multicolumn{2}{c|}{Prefix language Modeling, sequence split 75\%} \\
\hline
- & \multicolumn{2}{c|}{Next Token Prediction (also referred to as Causal language Modeling)(} \\
\hline
\end{tabular}
\caption{Summary of Denoising Objectives used in a specialized variant of UL2 for decoder-only models, as reported in \cite{garcia2023unreasonableeffectivenessfewshotlearning}. The objectives are sampled at a ratio of 20\% prefix language modeling, 60\% next token prediction, and 20\% infilling. Presumably, this translates to a 10\% sampling probability of the "R" dDenoiser and a 10\% sampling probability for the "S" denoiser. \textbf{Note:} a denoiser paradigm token was not provided for the next token prediction task, so. Thus, we do not include one here. More details are provided in \autoref{appendix:ul2_explanation}}
\label{appendix:tab:ul2_decoder_only_denoisers_table}
\end{table*}

UL2, introduced by Tay et al.~\cite{tay2023ul}, is a \textit{mixture-of-denoisers} (MoD) pre-training strategy originally developed and ablated for Transformer-based language models. Rather than relying solely on a causal (left-to-right) language modeling objective, UL2 interleaves multiple forms of denoising tasks, such as infilling and prefix language modeling, to produce more robust representations.

In its original formulation, UL2 labels each sequence with a “paradigm token” (e.g., “R” or “X”) to indicate which denoising objective was used. Table \ref{appendix:tab:ul2_denoisers_table} summarizes the full set of infilling and prefix modeling objectives in UL2, including their mean span widths and the percentage of tokens masked. These paradigm tokens were ablated off by continuing pre-training without them, for a significant number of steps.
~\citet{garcia2023unreasonableeffectivenessfewshotlearning} published an updated and specialized variant for decoder-only models from the authors of UL2. This variant augments UL2 with a pure next token prediction objective (standard causal language modeling) and simplifies the amount of infilling configurations. Table \ref{appendix:tab:ul2_decoder_only_denoisers_table} lists these modified denoisers, alongside their sampling ratios. In this variant, “R” denotes a short-span infilling objective, “X” denotes a more extreme infilling setup with higher mask percentages or larger spans, and “S” indicates a prefix language modeling task.

Empirical results in both the original and specialized versions show that mixing multiple denoising tasks can outperform training solely on next token prediction for many downstream tasks. In cases where UL2 does not surpass standard causal modeling, the authors noted performance differences tend to be small~\cite{tay2023ul}.

\subsection{Denoising Objectives and Birdie Motivation}
\label{appendix:ul2_denoisers_and_birdie}

Tables \ref{appendix:tab:ul2_denoisers_table} and \ref{appendix:tab:ul2_decoder_only_denoisers_table} provide an overview of the denoising strategies used in UL2. The objective categories, “R,” “X,” “S,” (as well as next token prediction), are used to group specific noising objective and configuration classes that are applied to input samples.
For instance, “R” often denotes standard span infilling at normal masking probabilities, while “X” can span larger regions of text and/or mask up to 50\% of tokens. In the case of “S” (prefix language modeling), the sequence is partially visible at the beginning, requiring the model to continue predicting from a given prefix.

In our approach with Birdie, we follow a similar paradigm-token strategy, but introduce two key modifications:
\begin{enumerate}
    \item \textbf{Appending Paradigm Tokens} Instead of always prepending the paradigm token, Birdie appends it 50\% of the time. This design choice is intended to familiarize the model with varied token placements and reduce reliance on position-based cues.
    \item \textbf{Dynamic Mixture Ratios.} While UL2 maintains fixed ratios of denoising tasks throughout training, Birdie adaptively updates the ratio of objectives (e.g., infilling vs. prefix vs. causal language modeling) based on model feedback signals. This avoids exhaustive ablation studies and aims to optimize performance for different architectures and tasks.
\end{enumerate}

These modifications address several limitations we observed when applying UL2 to \textit{state-space models} (SSMs):
\begin{enumerate}
    \item \textbf{Transformer-Centric Ablations.} UL2’s reported ablations focus primarily on Transformers. Our experiments indicate that SSM-based models (e.g., Gated SSMs) respond differently to certain denoising objectives, necessitating an approach tailored to SSMs.
    \item \textbf{Weak Retrieval/Copy Abilities.} Although UL2 excels on many downstream tasks, we found that SSMs trained with UL2 did not improve on exact-retrieval tasks (e.g., phone number copying) over a standard next-token predictor. This suggests the need for adaptive sampling of denoisers that more strongly emphasize copying skills for SSM architectures.
    \item \textbf{Fixed Objective Ratios.} UL2 uses a constant mixture of objectives across the entire training process. Empirically, we observed that certain objectives (e.g., infilling) may require a higher sampling frequency for SSMs, whereas Transformers are able to perform this task relatively easy. Dynamic ratio adjustments in Birdie allow the model to re-balance objectives on the fly, rather than relying on predefined proportions.
\end{enumerate}

Our approach is able to maintain the benefits of mixture-of-denoiser pre-training, while addressing our practical compute constraints and unique needs posed by SSMs.

\clearpage

\newgeometry{left=2.5cm,right=2.5cm,top=2cm,bottom=2cm}

\subsection{Hyperparameters}

\begin{table}[H]
\centering
\begin{threeparttable}
\small %
\begin{tabular}{|l|c|c|c|c|l|c|p{4cm}|}
\hline
\textbf{Model} & \textbf{HD} & \textbf{State} & \textbf{Block} & \textbf{MLP Exp.} & \textbf{Attn.} & \textbf{1D Conv} & \textbf{Bidir.} \\ \hline

Gated SSM & 2048 & 2560 & 48\tnote{*} & -- & -- & -- & Repeating: (50\% bidir., causal) \\ \hline

Hawk & 2048 & 2560 & 20 & $\frac{8}{3}$ & -- & Size 4 & Repeating: (50\% bidir., causal) \\ \hline

Transformer & 2048 & -- & 24 & $\frac{8}{3}$ & MHA 16H/128D & -- & Every layer \\ \hline
\end{tabular}
\begin{tablenotes}
\footnotesize
\item[*] Since Gated SSM uses a fused recurrence and MLP layer, similar parameter count is maintained.
\end{tablenotes}
\end{threeparttable}
\caption{Comparison of Model Characteristics}
\label{table:model_characteristics}
\end{table}

SSMs did not have weight decay applied to their $W^f$ weights, and Hawk does not have any weight decay applied to its RG-LRU parameters or biases.

\subsection{Phone Number Task}
\label{appendix:section:phone_number_task_and_sample}

\paragraph{Hyperparameters}
Models are fine-tuned for 1000 steps with no weight decay and a batch size of 64.
Training samples range from 200 to 800 entries and from 1-32 phone numbers to retrieve. Ideally, this allows for our models to handle any phone book example given in this range. We use sequence packing to concatenate shorter training examples out to 16384 tokens.
Hawk and Transformer models are trained with a fixed learning rate of $5 \times 10^{-5}$.

\paragraph{Worst Baselines Phonebook Sweep}

In an unsuccessful attempt to improve the non-converging models (Next Token Prediction, UL2, and Fixed Ratio Mixture), we ran extensive hyperparameter sweeps. Our best settings were a 0.0 global weight decay, 1e-5 learning rate, and a batch size of 64. Many settings achieved similar results, but others resulted in the accuracy collapsing to near 0\%.
Specifically, we tried the power set of the following hyperparameters:

\begin{itemize}
    \item \textbf{Global weight decay:} 0.0, 0.01, 0.1
    \item \textbf{Learning rate:} $1 \times 10^{-7}$, $5 \times 10^{-7}$, $1 \times 10^{-6}$, $5 \times 10^{-6}$, $1 \times 10^{-5}$, $5 \times 10^{-5}$, $1 \times 10^{-4}$, $5 \times 10^{-4}$, $1 \times 10^{-3}$
    \item \textbf{Batch size:} 16, 32, 64, 128, 256
\end{itemize}

\restoregeometry

\clearpage

\newgeometry{left=2.5cm,right=2.5cm,top=2cm,bottom=2cm}
\begin{figure*}[!ht]
\subsection{Birdie pre-training Metrics}
\centering
\includegraphics[angle=90,height=1.12\paperwidth,keepaspectratio]{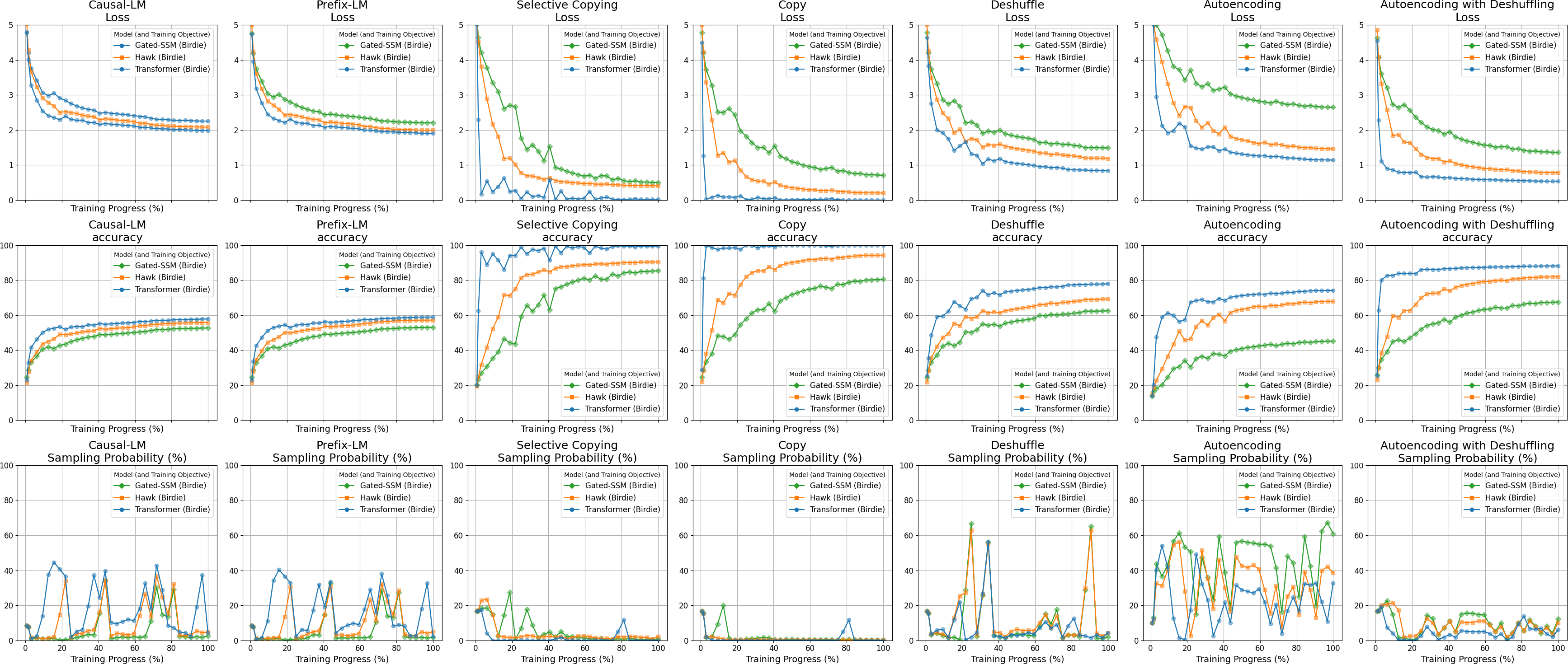}
\captionsetup{width=0.8\textwidth} 
\caption{These plots show how several metrics evolve during training. Loss and Accuracy are on Validation data from The Pile. Accuracy denotes greedy decoding accuracy. Sampling Probability (\%) denotes the probability that an objective in a class is selected for each segmented sample from the training dataloader, as selected by Birdie. The parameterizations for each objective are described in Appendix Section \ref{appendix:birdie_controls_overview}.}
\label{fig:APPENDIXRL}
\vspace*{0.08\paperheight} %
\end{figure*}
\restoregeometry

\clearpage

\subsection{Pre-training}
\label{subsec:Appendix-pre-training}

We train all models on the same data pipeline using The Pile \citep{gao2020pile}\footnote{We use the full version of The Pile, last available mid-2023}. The Pile is a collection of several datasets, and includes books, code, web scrapes, emails, and question-answer instruction formatted examples.

During all training and fine-tuning, we always use sequence packing and proper masking for all models, preventing samples from interfering with each other. For Hawk, we add spacing between samples to prevent the Conv1D layer from leaking information between samples. All models use this spacing to normalize the samples seen during evaluation periods and, therefore, reduce external noise when comparing models trained using Birdie's reinforcement learning setup. 

Models are trained for 32,000 steps, with a batch size of 520. We train all models on The Pile~\citep{gao2020pile} dataset for 32B tokens using sequence packing and proper masking to prevent sample interference. All models were pre-trained with a sequence length of 2048. Following recommendations by~\citet{chowdhery2022palm}, we pre-train slightly over Chinchilla optimal scaling laws~\citep{hoffmann2022training} -- 20-25x tokens per parameter. We provide a comparison of compute costs and resources in Table~\ref{tab:TrainingSpeeds}.
We count both context and target tokens as tokens "seen" by the model. This provides a fair comparison among different pre-training objectives. This diverges from other approaches, which do not always consider context tokens in their total count of tokens on which the model was trained~\citep{tay2023ul}. This means that the Copying task, for example, results in an actual reduction in the total count of unique training tokens seen by the model. This is because the training budget is for a number of tokens. With copying, the same tokens appear twice: once as an input, and once as a label.

We use the same hyperparameters for all models, using the same settings, such as learning rates and batch sizes, as models found in Mamba \cite{gu2023mamba}. We use the official settings for Hawk - specific gradient clipping on Beta (shown visible in appendix figure \ref{appendix:jax_code__hawk}), no weight decay on RG-LRU layers, and keep parameters stored in bfloat16 precision. All others use float32, though we always cast intermediates to bfloat16 except when running the recurrence functions for the Gated SSM and Hawk. Our Transformer baselines use a 250k decay rate for their RoPE positional encodings, following a suggestion on a Reddit post \cite{rozière2024codellamaopenfoundation}.

\subsection{Instruction Tuning}
\label{subsec:Appendix-Finetuning}

For 1.4B parameter models, we largely follow the progressive learning fine-tuning procedure from Orca 2~\citep{mitra2023orca}, as immediately jumping into relatively difficult, small datasets, such as SlimOrca-Dedup~\citep{slimorca-dedup} ended up hurting performance. We follow common instruction-tuning procedures from FLAN~\citep{longpre2023flan}, Zephyr~\citep{tunstall2023zephyr}, and Tulu~\citep{wang2023far} with dropout, cosine decay learning rate, and no weight decay.
We use all training, validation, and test sets as provided by the original authors.

We change hyperparameters from FLAN's paper since we use AdamW and not AdaFactor - we use a different learning rate to compensate for the lack of AdaFactor's parameter-scaled updates. We use a gentle 3e-4 peak cosine LR that decays down to 3e-5, similar to work in Zephyr \cite{tunstall2023zephyr} over 4 epochs. For FLAN, we extend the sequence length to 4096 (from 2048 during pre-training) and use a batch size of 20. This keeps the number of tokens per batch equal with the original publication. We finish instruction tuning by again resetting the optimizer state, and using a $3e-5$ to $3e-6 $cosine schedule over two epochs on the Open-Hermes dataset \cite{OpenHermes2.5}. During this final phase, we extend the sequence length to 8192, although the longest sample in Open-Hermes is only around 6,000 tokens long.

\newpage

\subsection{Hardware and Experimental Setup}
We train 11 models, each containing 1.4 billion parameters. The primary training is conducted over 5 machines, each equipped with 4 Nvidia A100 GPUs (80GB). Additionally, fine-tuning and evaluation was split among four NVIDIA H100 GPUs, five Google TPUv3-8, a TPUv4-32 slice, generously provided through Google's TPU Research Cloud, for which we are sincerely grateful.
The fixed ratios of the Fixed Ratio Mixture was found by training small 110M Gated SSM and Transformers models with random mixtures and hand-tuning sampling rates to ensure no objective. This took over 50 iterations of training the 110M model, which took roughly 5 hours each.

Table~\ref{tab:TrainingSpeeds} relates compute cost between models for the hardware we used for pre-training.

\begin{table}[htbp]
\centering
\scriptsize
\begin{tabular}{p{3em}p{7em}p{3em}p{2em}p{2.3em}p{3em}}
\toprule
\textbf{Backend} & \textbf{Model} & \textbf{GPU Hrs (A100)} & \textbf{Sec / Step} & \textbf{Seq Length} & \textbf{Tokens / sec / A100} \\ \midrule
Torch & Gated SSM & 3,200 & 2.0 & N/A & 26,148 \\ 
Torch & Flash Attn. 2 & 7,011 & 4.4 & 2048 & 12,152 \\ 
\midrule
JAX & Gated SSM & 5,600 & 3.5 & N/A & 15,214 \\ 
JAX & Hawk & 6,480 & 4.05 & N/A & 13,148 \\ %
JAX & Transformer & 10,016 & 6.3 & 2048 & 8,506 \\ \bottomrule
\end{tabular}
\caption{Comparison of observed model training speeds on our multi-node A100 setup.}
\label{tab:TrainingSpeeds}
\end{table}

\clearpage

\begin{table*}[htbp]
\subsection{The EleutherAI LM Harness Tasks}
\label{subsec:Appendix-EleutherAI}
\small
\centering
\begin{tabular}{|p{10em}|p{37em}|}
\hline
\textbf{Task} & \textbf{Description}\\\hline
ARC Easy & The 'easy' portion of a multiple-choice question-answering dataset, containing questions from science exams from grade 3 to 9~\citep{ARC-Challenge}.\\\hline
ARC Challenge & The Challenge portion of the dataset, containing the more difficult questions that require reasoning~\citep{ARC-Challenge}.\\\hline
BoolQ & A question answering dataset for Yes/No questions containing 15942 examples; each example is a triplet of (question, passage, answer), with the title of the page (from google search engine where questions are collected) as optional additional context~\citep{clark-etal-2019-boolq}. \\\hline
COPA &  The Choice Of Plausible Alternatives (COPA) dataset consists of 1000 questions composed of a premise and two alternatives, with the task being to select the alternative that more plausibly has a causal relation with the premise~\citep{gordon-etal-2012-semeval}.	\\ \hline
HellaSwag & A dataset designed to test common sense reasoning and grounded situations, presenting contexts from video and text with multiple-choice endings where a model must predict the most likely continuation~\citep{zellers2019hellaswag}.\\\hline
LogiQA & A question answering dataset derived from logical reasoning examination questions, aimed at evaluating the deep logical reasoning capability of models~\citep{liu2020logiqa}.\\\hline
MathQA & A large-scale dataset of math word problems~\citep{amini-etal-2019-mathqa}. \\\hline
MC-TACO & 13K question-answer pairs that require temporal commonsense comprehension on (1) duration of an event, (2) order of events, (3) time when event occurs, (4) event frequency, and (5) stationarity (whether a state is maintained for a very long time or indefinitely).~\citep{zhou-etal-2019-going}   \\	\hline
MedMCQA & A large-scale, Multiple-Choice Question Answering (MCQA) dataset designed to address real-world medical entrance exam questions~\citep{pal2022medmcqa}.\\\hline	
MMLU & The Massive Multitask Language Understanding (MMLU) dataset, consisting of questions spanning multiple subjects and domains, designed to test models on a broad range of knowledge and reasoning skills~\citep{hendryckstest2021}.\\\hline
MNLI & Often also referred to as multi-nl, this Multi-Genre Natural Language Inference (MultiNLI) corpus is a dataset to test sentence understanding; it offers data from ten distinct genres of written and spoken English--enabling evaluation on nearly the full complexity of the language and on cross-genre domain adaptation.~\citep{williams-etal-2018-broad} \\	\hline
OpenBookQA & A dataset that consists of 5,957 multiple-choice questions that necessitate the use of both reasoning and additional broad common sense or scientific knowledge not contained in the question itself~\citep{OpenBookQA2018}.\\\hline
PIQA & The Physical Interaction Question Answering dataset, focusing on reasoning about physical properties of objects and the actions taken upon them~\citep{Bisk2020}.\\\hline
PubMedQA & A Yes/No biomedical question answering dataset collected from PubMed abstracts~\citep{jin2019pubmedqa}.\\	\hline
qa4mre & The Question Answering for Machine Reading Evaluation dataset is designed for the annual competition, consisting of a series of questions based on a single document with multiple-choice answers~\citep{Peas2013QA4MRE2O}.\\ \hline
QNLI & The Question-answering Natural Language Inference dataset is automatically derived from the Stanford Question Answering Dataset v1.1 (SQuAD) of question-paragraph pairs, where one of the sentences in the paragraph (drawn from Wikipedia) contains the answer to the corresponding question (written by an annotator).~\citep{wang-etal-2018-glue}.  \\	\hline
race & A large-scale reading comprehension dataset collected from English exams, featuring questions with multiple-choice answers that demand high-level reasoning abilities~\citep{lei2018simple}.\\\hline
SciQ & Crowd-sourced science exam questions about Physics, Chemistry, Biology, etc, in multiple-choice format with 4 answer options and an evidence-supporting paragraph for the correct answer for most questions~\citep{welbl-etal-2017-crowdsourcing}. \\ \hline
SST-2 & The Stanford Sentiment Treebank, a corpus with fully labeled parse trees for a complete analysis of the compositional effects of sentiment in language~\citep{socher-etal-2013-recursive}.\\	\hline
WiC & A large-scale Word in Context dataset based on annotations curated by experts for generic evaluation of context-sensitive representations~\citep{WIC-Task}.\\ \hline
Winogrande & A large-scale dataset of 44k problems, inspired by the original Winograd Schema Challenge (WSC) design~\citep{WSC}, but adjusted to improve both the scale and the hardness of the dataset~\citep{Winogrande}.\\
\hline
\end{tabular}
\end{table*}

\newpage
\clearpage

\newpage

\clearpage

\begin{sidewaystable}[!ht]
    \centering
    \resizebox{\textwidth}{!}{
    \begin{tabular}{|l|l|c|c|c|c|c|c|c|c|c|c|c|c|c|c|c|c|c|c|c|c|c|c|c|}
        \hline
        \multicolumn{25}{|l|}{\textbf{Instruct Models}} \\
        \hline
        \textbf{Model} & \textbf{Procedure} & \textbf{Average} & \textbf{ARC Challenge} & \textbf{ARC Easy} & \textbf{BoolQ} & \textbf{COPA} & \textbf{HellaSwag} & \textbf{LogiQA} & \textbf{MathQA} & \textbf{MC-TACO} & \textbf{MedMCQA} & \textbf{MMLU} & \textbf{MNLI} & \textbf{OpenBookQA} & \textbf{PIQA} & \textbf{PubMedQA} & \textbf{QA4MRE} & \textbf{QNLI} & \textbf{race} & \textbf{SciQ} & \textbf{SST-2} & \textbf{TruthfulQA (MC1)} & \textbf{WiC} & \textbf{Winogrande}  \\
        \hline
        Hawk & Birdie & 41.4\% & 25.4\% & 31.8\% & 61.5\% & 43.8\% & 30.0\% & 31.6\% & 26.2\% & 63.7\% & 29.8\% & 22.1\% & 32.3\% & 29.4\% & 62.9\% & 53.6\% & 31.4\% & 57.3\% & 29.8\% & 43.6\% & 74.9\% & 28.8\% & 49.8\% & 51.1\%  \\ \hline
        Hawk & Birdie - Causal & 40.8\% & 26.3\% & 32.3\% & 52.9\% & 45.4\% & 30.6\% & 31.3\% & 25.8\% & 55.4\% & 28.3\% & 22.9\% & 31.8\% & 31.2\% & 63.5\% & 48.6\% & 28.5\% & 55.2\% & 27.7\% & 43.8\% & 81.5\% & 30.8\% & 53.6\% & 49.9\%  \\ \hline
        Hawk & Next Token Pred & 39.6\% & 25.7\% & 32.5\% & 62.1\% & 46.9\% & 28.5\% & 27.6\% & 26.5\% & 62.1\% & 26.9\% & 22.9\% & 33.2\% & 31.0\% & 60.9\% & 52.2\% & 25.0\% & 49.9\% & 27.6\% & 45.2\% & 54.9\% & 27.5\% & 50.3\% & 50.7\%  \\ \hline
        Attention & Birdie & 39.7\% & 25.3\% & 30.7\% & 62.5\% & 45.7\% & 31.4\% & 30.6\% & 25.7\% & 36.0\% & 31.7\% & 22.6\% & 31.8\% & 32.8\% & 62.2\% & 54.4\% & 33.9\% & 49.8\% & 29.6\% & 54.2\% & 50.9\% & 29.1\% & 50.0\% & 51.9\%  \\ \hline
        Attention & Next Token Pred & 40.4\% & 26.5\% & 33.0\% & 50.1\% & 46.6\% & 33.8\% & 32.0\% & 26.7\% & 41.0\% & 25.8\% & 26.7\% & 31.8\% & 30.8\% & 63.9\% & 47.8\% & 32.8\% & 51.0\% & 30.2\% & 42.7\% & 87.6\% & 26.2\% & 50.6\% & 50.9\% \\ \hline
        \hline
        \multicolumn{25}{|l|}{\textbf{Base Models}} \\
        \hline
        \textbf{Model} & \textbf{Procedure} & \textbf{Average} & \textbf{ARC Challenge} & \textbf{ARC Easy} & \textbf{BoolQ} & \textbf{COPA} & \textbf{HellaSwag} & \textbf{LogiQA} & \textbf{MathQA} & \textbf{MC-TACO} & \textbf{MedMCQA} & \textbf{MMLU} & \textbf{MNLI} & \textbf{OpenBookQA} & \textbf{PIQA} & \textbf{PubMedQA} & \textbf{QA4MRE} & \textbf{QNLI} & \textbf{race} & \textbf{SciQ} & \textbf{SST-2} & \textbf{TruthfulQA (MC1)} & \textbf{WiC} & \textbf{Winogrande}  \\
        \hline
        Hawk & Birdie & 38.3\% & 25.9\% & 33.2\% & 42.3\% & 48.1\% & 32.3\% & 26.3\% & 24.8\% & 53.3\% & 22.8\% & 28.0\% & 31.8\% & 31.0\% & 62.9\% & 36.8\% & 26.4\% & 50.2\% & 27.5\% & 50.1\% & 58.3\% & 29.5\% & 50.9\% & 49.8\%  \\ \hline
        Hawk & Birdie - Causal & 39.0\% & 25.0\% & 34.6\% & 49.3\% & 45.9\% & 32.7\% & 30.0\% & 25.6\% & 40.9\% & 24.3\% & 26.0\% & 31.8\% & 29.4\% & 63.3\% & 48.2\% & 24.8\% & 49.4\% & 28.1\% & 60.0\% & 61.0\% & 28.0\% & 49.8\% & 50.5\%  \\ \hline
        Hawk & Next Token Pred & 39.1\% & 25.4\% & 34.6\% & 55.4\% & 49.7\% & 34.6\% & 26.6\% & 25.1\% & 35.1\% & 29.9\% & 23.6\% & 32.0\% & 30.8\% & 61.8\% & 55.2\% & 29.1\% & 48.5\% & 26.7\% & 54.6\% & 55.5\% & 25.1\% & 49.4\% & 51.9\%  \\ \hline
        Attention & Birdie & 38.5\% & 23.0\% & 28.8\% & 44.3\% & 48.4\% & 29.7\% & 33.0\% & 24.7\% & 66.1\% & 28.1\% & 25.3\% & 31.8\% & 22.2\% & 55.3\% & 37.2\% & 30.0\% & 51.7\% & 29.3\% & 62.5\% & 50.0\% & 27.7\% & 48.9\% & 49.6\%  \\ \hline
        Attention & Next Token Pred & 39.9\% & 26.0\% & 35.0\% & 62.1\% & 50.4\% & 40.1\% & 31.0\% & 26.4\% & 33.9\% & 24.1\% & 26.5\% & 31.8\% & 31.8\% & 65.5\% & 55.2\% & 25.5\% & 49.4\% & 30.8\% & 55.6\% & 50.0\% & 25.9\% & 50.0\% & 50.7\% \\ \hline
        \hline
    \end{tabular}
    }
    \caption{Hawk and Transformer model performance comparisons for Instruct Models and Base Models.}
\end{sidewaystable}

\begin{sidewaystable}[!ht]
    \centering
    \resizebox{\textwidth}{!}{
    \begin{tabular}{|l|l|c|c|c|c|c|c|c|c|c|c|c|c|c|c|c|c|c|c|c|c|c|c|c|}
        \hline
        \multicolumn{25}{|l|}{\textbf{Instruct Models}} \\
        \hline
        \textbf{Model} & \textbf{Procedure} & \textbf{Average} & \textbf{ARC Challenge} & \textbf{ARC Easy} & \textbf{BoolQ} & \textbf{COPA} & \textbf{HellaSwag} & \textbf{LogiQA} & \textbf{MathQA} & \textbf{MC-TACO} & \textbf{MedMCQA} & \textbf{MMLU} & \textbf{MNLI} & \textbf{OpenBookQA} & \textbf{PIQA} & \textbf{PubMedQA} & \textbf{QA4MRE} & \textbf{QNLI} & \textbf{race} & \textbf{SciQ} & \textbf{SST-2} & \textbf{TruthfulQA (MC1)} & \textbf{WiC} & \textbf{Winogrande}  \\
        \hline
        Gated SSM & Birdie & 41.2\% & 25.6\% & 31.4\% & 58.7\% & 47.6\% & 29.0\% & 28.6\% & 25.3\% & 64.7\% & 31.7\% & 21.6\% & 31.8\% & 30.4\% & 60.6\% & 54.2\% & 28.9\% & 53.4\% & 27.9\% & 43.9\% & 82.1\% & 28.2\% & 51.3\% & 50.0\%  \\ \hline
        Gated SSM & Birdie - Causal & 41.1\% & 25.9\% & 32.4\% & 52.0\% & 45.8\% & 28.8\% & 26.1\% & 25.7\% & 65.3\% & 32.2\% & 21.7\% & 31.9\% & 31.0\% & 62.2\% & 52.6\% & 28.2\% & 50.9\% & 29.0\% & 43.7\% & 87.5\% & 30.8\% & 50.0\% & 50.2\%  \\ \hline
        Gated SSM & Fixed Ratio Mix & 38.8\% & 25.7\% & 32.7\% & 61.4\% & 43.3\% & 29.5\% & 29.0\% & 25.2\% & 42.5\% & 32.0\% & 21.2\% & 31.8\% & 30.2\% & 62.3\% & 23.2\% & 25.9\% & 53.2\% & 29.4\% & 42.0\% & 77.6\% & 29.3\% & 52.4\% & 52.6\%  \\ \hline
        Gated SSM & Next Token Pred & 38.7\% & 25.6\% & 32.4\% & 61.5\% & 47.7\% & 31.2\% & 27.0\% & 25.8\% & 34.1\% & 28.5\% & 21.7\% & 31.8\% & 30.6\% & 61.9\% & 54.6\% & 27.8\% & 49.5\% & 28.5\% & 43.6\% & 54.5\% & 31.9\% & 50.2\% & 49.8\%  \\ \hline
        Gated SSM & UL2 & 39.7\% & 25.3\% & 31.7\% & 52.3\% & 45.9\% & 28.6\% & 26.3\% & 25.1\% & 62.1\% & 30.5\% & 21.5\% & 31.8\% & 31.0\% & 61.5\% & 44.8\% & 29.4\% & 48.8\% & 26.0\% & 40.8\% & 79.8\% & 29.7\% & 49.2\% & 50.4\% \\ \hline
        \multicolumn{25}{|l|}{\textbf{Base Models}} \\
        \hline
        \textbf{Model} & \textbf{Procedure} & \textbf{Average} & \textbf{ARC Challenge} & \textbf{ARC Easy} & \textbf{BoolQ} & \textbf{COPA} & \textbf{HellaSwag} & \textbf{LogiQA} & \textbf{MathQA} & \textbf{MC-TACO} & \textbf{MedMCQA} & \textbf{MMLU} & \textbf{MNLI} & \textbf{OpenBookQA} & \textbf{PIQA} & \textbf{PubMedQA} & \textbf{QA4MRE} & \textbf{QNLI} & \textbf{race} & \textbf{SciQ} & \textbf{SST-2} & \textbf{TruthfulQA (MC1)} & \textbf{WiC} & \textbf{Winogrande}  \\
        \hline
        Gated SSM & Birdie & 37.2\% & 25.0\% & 30.2\% & 38.8\% & 46.5\% & 28.6\% & 26.1\% & 24.3\% & 62.8\% & 21.6\% & 23.5\% & 31.8\% & 30.4\% & 60.4\% & 34.4\% & 26.1\% & 52.1\% & 25.1\% & 42.1\% & 57.9\% & 29.7\% & 50.0\% & 50.1\%  \\ \hline
        Gated SSM & Birdie - Causal & 37.5\% & 24.7\% & 27.9\% & 49.9\% & 43.7\% & 28.5\% & 25.8\% & 24.4\% & 47.2\% & 28.8\% & 21.9\% & 32.1\% & 31.8\% & 61.2\% & 50.0\% & 24.1\% & 50.4\% & 24.6\% & 41.4\% & 53.7\% & 31.5\% & 50.0\% & 51.7\%  \\ \hline
        Gated SSM & Fixed Ratio Mix & 38.9\% & 25.1\% & 30.9\% & 39.1\% & 47.3\% & 29.6\% & 27.0\% & 24.0\% & 66.1\% & 26.9\% & 23.1\% & 31.8\% & 30.6\% & 59.6\% & 53.6\% & 28.7\% & 50.6\% & 27.4\% & 46.5\% & 59.2\% & 27.7\% & 50.0\% & 51.0\%  \\ \hline
        Gated SSM & Next Token Pred & 39.1\% & 25.5\% & 34.1\% & 62.0\% & 48.5\% & 35.5\% & 27.2\% & 24.3\% & 34.1\% & 32.2\% & 21.2\% & 31.8\% & 30.2\% & 64.6\% & 52.0\% & 27.5\% & 49.3\% & 29.3\% & 48.9\% & 56.9\% & 25.5\% & 50.0\% & 49.6\%  \\ \hline
        Gated SSM & UL2 & 38.0\% & 23.6\% & 33.9\% & 43.9\% & 44.2\% & 28.8\% & 25.5\% & 24.5\% & 53.3\% & 30.0\% & 22.9\% & 31.8\% & 32.2\% & 61.2\% & 40.8\% & 25.5\% & 49.8\% & 27.3\% & 49.9\% & 58.7\% & 27.1\% & 50.8\% & 50.8\% \\ \hline
    \end{tabular}
    }
    \caption{Gated SSM training procedure comparisons performance comparisons for Instruct Models and Base Models.}
    \label{tab:PerformancePerEleutherAITask}
\end{sidewaystable}

\clearpage
\newpage

\subsection{SQuAD V2: Question and Answering}
\label{appendix:section:squad_v2}

\paragraph{Task Description and Setup}

We evaluate our instruction-tuned models on SQuAD V2, a question-answering dataset where models are provided with a Wikipedia excerpt and asked a question. Some questions have multiple acceptable answers, while others are intentionally unanswerable. Following previous work \citep{jelassi2024repeat}, we focus exclusively on answerable questions and do not fine-tune our models on this task.

The standard metric for SQuAD V2 (F1) penalizes models for verbosity by reducing scores when additional words are present. Since our models are not trained to prioritize brevity, and SQuAD V2 predates today’s conversational language models, we place greater emphasis on the "Answer Contains Label" metric. This metric awards full credit if any acceptable answer fully appears in the model's response, whereas the F1 score grants partial credit for matching words but penalizes longer responses.

\begin{table}[htbp]
\centering
\scriptsize 
\begin{tabular}{@{}lccc@{}}
\toprule
\textbf{Model Tag} & \textbf{Training Procedure} & \textbf{F1 (\%)} & \textbf{Answer Contains Label (\%)} \\
\midrule
\textbf{Transformer}         & \textbf{Birdie}                     & 21.4 & \textbf{73.7} \\
Transformer         & Next Token Prediction                        & 21.0 & 60.9 \\
\textbf{Hawk}          & \textbf{Birdie}                     & \textbf{23.2} & 54.4 \\
Hawk          & Next Token Prediction                        & 9.1  & 15.7 \\
\bottomrule
\end{tabular}
\caption{Averaged SQuAD V2 results with instruction-tuned models. Training with the Birdie procedure strongly improves SSM performance, compared to Next Token Prediction. The best performing models and metrics are shown in bold. These results are plotted by sequence length in Figure \ref{fig:squad}}.
\label{tab:squadv2_hawk_transformer_model_performance}
\end{table}

\begin{table}[htbp]
\centering
\scriptsize 
\begin{tabular}{@{}lccc@{}}
\toprule
\textbf{Model Tag} & \textbf{Training Procedure} & \textbf{F1 (\%)} & \textbf{Answer Contains Label (\%)} \\
\midrule
\textbf{Gated SSM} & \textbf{Birdie}                     & \textbf{17.0} & \textbf{31.3} \\
Gated SSM           & UL2                        & 12.8 & 18.6 \\
Gated SSM           & Fixed Ratio Mixture       & 11.3 & 18.5 \\
Gated SSM           & Birdie - Causal            & 11.3 & 15.0 \\
Gated SSM           & Next Token Prediction                        & 10.3 & 14.7 \\
\bottomrule
\end{tabular}
\caption{Averaged SQuAD V2 Question-Answering results with instruction-tuned Gated SSM models. Training with the Birdie procedure strongly improves SSM performance compared to other training procedures. The best performing model and metrics are shown in bold.}
\label{tab:squadv2_ablation_model_performance}
\end{table}

\clearpage

\subsection{Story Infilling Task}
\label{appendix:section:infilling_task_info}

\paragraph{Task Description}
We generate thousands of stories with blank sections using Mistral v0.1 Instruct (7B) with an unusually high temperature of 10.0 and use a $min_p$ of 0.10 to keep text coherent. At the same time, we have the model generate four potential choices to fill in that story, with one of them being the intended best choice.
Generally, the choices to fill in the stories are plausible. The model tends to generate at least one adversarial option that is very close to being the best answer, but is also not the best choice.

We filter questions using a Jaccard similarity of 0.85, so when at least two stories share at least 15\% of their words, only one is kept and the rest are removed.
Finally, we present each story and its choices to four language models, and ask if the intended label is truly the best choice. We remove questions that do not receive a majority vote from four language models. Specifically, these are the instruct versions of Mistral Nemo 2407 (12B), Gemma-2 (9B), Llama 3.1 (8B), and Mistral v0.3 (7B).

\begin{table}[h!]
\centering
\scriptsize 
\begin{tabular}{p{7em}lc}
\toprule
\textbf{Model} & \textbf{Training Procedure} & \textbf{Accuracy} \\ 
\midrule
\multicolumn{3}{l}{\textbf{Instruct Models}} \\
Gated SSM & Birdie & 36.8\% \\
Gated SSM & Fixed Ratio Mixture & 36.2\% \\
Gated SSM & UL2 & 34.7\% \\
Gated SSM & Birdie - Causal & 33.9\% \\
Gated SSM & Next Token Prediction & 32.2\% \\
\midrule
\multicolumn{3}{l}{\textbf{Base Models}} \\
Gated SSM & Birdie & 36.8\% \\
Gated SSM & Birdie - Causal & 34.7\% \\
Gated SSM & UL2 & 31.7\% \\
Gated SSM & Fixed Ratio Mixture & 29.6\% \\
Gated SSM & Next Token Prediction & 27.5\% \\
\bottomrule
\end{tabular}
\caption{Average accuracy over our new infilling dataset. Models fill in a missing part of a story by selecting the best possible option. Losses are normalized by target token length.}
\label{appendix:infilling_ablation_result_table}
\end{table}

\clearpage
\paragraph{Dataset Example}
Below, we provide three figures for the infilling task. Figure \ref{appendix_infilling_dataset_example_short} shows a shorter entry in the dataset, and Figure \ref{appendix:infilling_long_example} shows the longest entry from our new infilling dataset. Finally, we show the prompt given to language models to judge the validity of questions and labels.

\begin{figure}[H]
    \begin{tcolorbox}[colframe=black, colback=white, boxrule=1pt, arc=4mm, width=2.0\columnwidth]
        \centering
    Short Entry:
    \begin{lstlisting}[style=phoneBookExample]
    Consider the following sequence of events, then select a choice that best fills in the missing entry:
    1. A stranger hands a letter to Ellie on a rainy afternoon.
    2. (blank)
    3. As she gets closer to the island, the edges of the map feel warm.
    
    Choices:
    (A) The letter contains information about a secret meeting happening at the end of the week.
    (B) She ignores the letter and throws it away.
    (C) Ellie finds a hidden treasure map in the envelope.
    (D) The letter leads her to an uncharted island.
    
    Which choice best fills in the missing entry?
    \end{lstlisting}
    
    Label:
    \begin{lstlisting}[style=phoneBookExample]
    (D) The letter leads her to an uncharted island.
    \end{lstlisting}
    \end{tcolorbox}
    
    \caption{A short example from our new infilling task. For more details, see appendix section \ref{appendix:section:infilling_task_info}.}
    \label{appendix_infilling_dataset_example_short}
\end{figure}

\begin{figure*}[h!]
    \begin{tcolorbox}[colframe=black, colback=white, boxrule=1pt, arc=4mm, width=2.0\columnwidth]
        Long Entry:
        \begin{lstlisting}[style=phoneBookExample]
            Consider the following sequence of events, then select a choice that best fills in the missing entry:
            1. A young woman named Mia had a passion for baking. She enjoyed trying out new recipes and experimenting with different flavors. One day, as she was perusing through a cookbook, she came across a recipe for a unique chocolate cake that sounded both delicious and challenging to make. Determined to create this masterpiece, Mia gathered all the necessary ingredients and began the process.
            2. (blank)
            3. She added more flour to thicken the mixture and waited patiently for the result. When she took a small spoonful of the new mixture, it had finally reached a consistency that resembled cake batter. Relieved, Mia continued with her baking process, pouring the mixture into a round pan and placing it in the oven.
            4. The aroma of freshly baked chocolate cake filled Mia's home as she waited for the timer to go off. When the cake was finished, she carefully removed it from the pan and placed it on a cooling rack. Once it had cooled down enough to eat, Mia took a bite and smiled with satisfaction. Her experimentation had paid off; she had created a delectable chocolate cake that tasted as good as it smelled.
            5. Proud of her achievement, Mia shared the cake with her family. They all raved about how moist and flavorful the cake was, with no one guessing the troubles she had gone through to perfect the recipe. From that day on, this new chocolate cake recipe became a staple in Mia's kitchen, something that both delighted her family and showcased her unwavering determination to succeed in all things baking.
            
            Choices:
            (A) The chocolate cake mixture seemed too watery, so Mia added an additional ingredient.
            (B) Mia decided that she did not need to adjust the recipe and proceeded with it as written.
            (C) Mia gave up on her goal of creating the perfect chocolate cake.
            (D) Mia added more flour to thicken the mixture.
            
            Which choice best fills in the missing entry?
        \end{lstlisting}
        
        Label:
        \begin{lstlisting}[style=phoneBookExample]
            (A) The chocolate cake mixture seemed too watery, so Mia added an additional ingredient.
        \end{lstlisting}
    \end{tcolorbox}
    \caption{A long example from our new infilling task. For more details, see appendix section \ref{appendix:section:infilling_task_info}.}
    \label{appendix:infilling_long_example}
\end{figure*}

\begin{figure*}[h!]
    \begin{tcolorbox}[colframe=black, colback=white, boxrule=1pt, arc=4mm, width=2.0\columnwidth]
    
        \begin{lstlisting}[style=phoneBookExample]
We are making a dataset. Please help us determine if the possible label is the best available choice from the given options.
Given a sequence of events with a missing entry, models are supposed to predict the best choice to fill in the blank line.

Here is the sequence of events:
1. A group of scientists embark on a mission to explore the outer edges of the universe.
2. (blank)
3. Their discoveries revolutionize our understanding of cosmic phenomena.

Here are the possible choices:
(A) The scientists fail to return to Earth due to an unknown celestial event.
(B) They discover a new planet that can sustain human life.
(C) They encounter another species of intelligent beings from beyond our galaxy.
(D) They uncover the secrets behind black holes and how they function.

Is "(D) They uncover the secrets behind black holes and how they function." the best choice out of the above for replacing the blank line?

Reply immediately with yes or no.
\end{lstlisting}

    \end{tcolorbox}
    \caption{The prompt used to ask models if the question and label are suitable for the infilling task. For more details, see appendix section \ref{appendix:section:infilling_task_info}.}
    \label{appendix:infilling_prompt_to_judges}
\end{figure*}

\clearpage

\FloatBarrier

\begin{figure*}[h!]
\begin{tcolorbox}[colframe=black, colback=white, boxrule=1pt, arc=4mm, width=2.0\columnwidth]

Inputs:
\begin{lstlisting}[style=phoneBookExample]
What are the phone numbers for Keven Meador, Stacey Krohn, Aubrey Wrenn, Eva Jurkovic, Gloria Job, Lamont Wilson, Emerald Hyman, Ali Hunsberger, Karsyn Jankowski, Alec Vinyard, Cole Pattison, Noe Pacheco, Trent Adamo, Greggory Chudnovsky, Yandel Funderburk, Scot Mitterer, Matthew Zeigler, Delvin Lerdal, Ellen Hickerson, Violet Lightbody, Ashlynn Buckingham, Pranav Blaisdell, Sheridan Lorentz, Levar Sharpe, Ramiro Vanlandingham, Yahir Leavitt, Cassius Mcguigan, Lillie Jetmore, Beatriz Jobe, Jamison Arruda, John Lovett, and Wade Anger? Find them in the phonebook below.

Phonebook:
Leonardo Rampone: 669-174-4914
Porter Wendell: 243-610-6940
Nicolle Journell: 612-425-4786
Tremayne Wcislo: 811-843-0927
[[~12 pages worth of phone entries go here]]
Elbert Foglesong: 345-541-6086
Matthew Zeigler: 417-648-0710
Patricia Queener: 174-489-9656
Kathryn Enrile: 472-553-8622

What are the phone numbers for Keven Meador, Stacey Krohn, Aubrey Wrenn, Eva Jurkovic, Gloria Job, Lamont Wilson, Emerald Hyman, Ali Hunsberger, Karsyn Jankowski, Alec Vinyard, Cole Pattison, Noe Pacheco, Trent Adamo, Greggory Chudnovsky, Yandel Funderburk, Scot Mitterer, Matthew Zeigler, Delvin Lerdal, Ellen Hickerson, Violet Lightbody, Ashlynn Buckingham, Pranav Blaisdell, Sheridan Lorentz, Levar Sharpe, Ramiro Vanlandingham, Yahir Leavitt, Cassius Mcguigan, Lillie Jetmore, Beatriz Jobe, Jamison Arruda, John Lovett, and Wade Anger? Find them in the phonebook above.
\end{lstlisting}

Labels:
\begin{lstlisting}[style=phoneBookExample]
337-743-1822, 487-090-9300, 261-549-5474, 239-751-7415, 899-328-4576, 500-199-0084, 744-974-9713, 617-979-7448, 132-114-9918, 807-843-6708, 200-177-4367, 800-256-6603, 276-090-4864, 174-449-8065, 107-912-1144, 367-994-8279, 417-648-0710, 130-012-0838, 668-436-3798, 951-625-4252, 734-538-6288, 952-422-8127, 209-140-8566, 252-088-9435, 956-578-5675, 355-111-4554, 779-940-5640, 235-150-3054, 312-638-2822, 400-177-6943, 896-686-1785, 330-123-2864
\end{lstlisting}
\label{appendix:abbreviated_phonebook_example}
\end{tcolorbox}
\caption{An abbreviated example of a 32 phone number retrieval sample with a 16,384 token length.}
\end{figure*}

\newpage
\clearpage

\subsubsection{Gated SSM Implementation}
\label{appendix:jax_code__gated_linear_rnn}

\begin{lstlisting}[language=Python,style=pythonCode, breaklines=true]
import jax
import jax.numpy as jnp
from jax.nn import sigmoid, gelu
from flax.linen import Module, Dense

class GatedLinearRNN(Module):
  state_size: int
  hidden_size: int

  def setup(self):
    self.W_f = Dense(self.state_size)
    self.W_z_gate = Dense(self.state_size)
    self.W_z = Dense(self.state_size)
    self.W_out_gate = Dense(self.state_size)
    self.W_out = Dense(self.hidden_size)


  def __call__(self, x_t):
    out_gate = gelu(self.W_out_gate(x_t))
    
    f_t = sigmoid(self.W_f(x_t))
    z_t = self.W_z(x_t) *
        sigmoid(self.W_z_gate(x_t))
    
    h_t = ParallelScan(f_t, z_t)
    y_t = self.W_out(out_gate * h_t)
    return y_t
    
\end{lstlisting}

\newpage
\clearpage

\subsubsection{Hawk Implementation}
\label{appendix:jax_code__hawk}
\begin{lstlisting}[language=Python,style=pythonCode, breaklines=true]
import jax
import jax.numpy as jnp
from jax.nn import sigmoid, softplus
from jax import custom_vjp
import flax.linen as nn
from flax.linen import Module, Dense



""" Hawk is untrainable without aggressive gradient clipping (standard gradient norm clipping is insufficient).
This custom backwards pass implementation is directly from RG-LRU code in the RecurrentGemma codebase. """

@custom_vjp
def sqrt_bound_derivative(x, max_gradient):
    """ Computes a square root with a gradient clipped at 'max_gradient'. """
    return jnp.sqrt(x)

def stable_sqrt_fwd(x, max_gradient):
    return jnp.sqrt(x), (x, max_gradient)

def stable_sqrt_bwd(res, g):
    x, max_gradient = res
    x_clipped = jnp.maximum(x, 1 / (4 * max_gradient**2))
    return (g / (2 * jnp.sqrt(x_clipped)),)

sqrt_bound_derivative.defvjp(stable_sqrt_fwd, stable_sqrt_bwd)


class HawkLayer(nn.Module):
    """Hawk Layer: This layer uses a Conv1D followed by an RG-LRU layer.
    
    Attributes:
        forget_base: Base forgetting factor.
        alpha_log_scale: "C" in the RG-LRU equation. Scaling factor for the alpha parameter.
        max_gradient: Maximum gradient for (NaN) gradient clipping in sqrt operation.
    """
    forget_base: float
    alpha_log_scale: float
    state_size: int
    d_model: int
    max_gradient: float = 1000.0

    def setup(self):
        self.W_a = Dense(self.state_size)
        self.W_x = Dense(self.state_size)
        self.W_input = Dense(self.state_size, use_bias=False)
        self.W_output = Dense(self.d_model, use_bias=False)
        self.W_gate = Dense(self.state_size, use_bias=False)
        self.Conv1D = Conv(features=state_size, kernel_size=4)


    def __call__(self, x_t):
        sidegate = gelu(self.W_gate(x_t))
        x_t = self.Conv1D(x_t)
        
        r_t = sigmoid(self.W_a(x_t))
        softplus_forget_base = softplus(self.forget_base)
        
        a_t = jnp.exp(self.alpha_log_scale * softplus_forget_base * r_t)
        log_a = -8.0 * gate_a * jax.nn.softplus(a_param)
        a = jnp.exp(log_a)

        
        a_squared = jnp.exp(2 * log_a)
        beta = sqrt_bound_derivative(
                    1 - a_squared,
                    self.max_gradient
                )
        i_t = (beta * sigmoid(self.W_x(x_t)) * x_t)
        
        h_t = ParallelScan(a_t, i_t)
        y_t = self.W_output(sidegate * h_t)
        return y_t

\end{lstlisting}

\end{document}